\documentclass[runningheads]{llncs}

\usepackage{arxiv_style_2026/arxiv}
\usepackage[width=122mm,left=12mm,paperwidth=146mm,height=193mm,top=12mm,paperheight=217mm]{geometry}

\usepackage{arxiv_style_2026/arxivabbrv}

\usepackage{graphicx}
\usepackage{booktabs}
\usepackage{bigdelim}
\usepackage{multirow}
\usepackage[accsupp]{axessibility}  %

\usepackage{hyperref}

\usepackage{orcidlink}

\usepackage{multirow}
\usepackage{comment}

\newcommand{\suppl}{\textcolor{black}{suppl. material}\xspace}

\renewcommand{\footnotesize}{\fontsize{8pt}{9.6pt}\selectfont}
\newcommand\samethanks[1][\value{footnote}]{\footnotemark[#1]}

\begin{document}

\title{Head Avatars with Dynamic Explicit Hair} 

\titlerunning{Head Avatars with Dynamic Explicit Hair}

\author{Vanessa Sklyarova\inst{1,2}\thanks{Equal contribution. \quad $^{\dagger}$ Equal supervision.}\orcidlink{0000-0002-8883-9972} \and
Haonan Chen\inst{1}\samethanks\orcidlink{0009-0008-4597-4016} \and
Berna Kabadayi\inst{2,4}\orcidlink{0009-0000-2882-8561} \and
Tobias Kirschstein\inst{5}\orcidlink{0009-0002-5308-591X} \and
Zicong Fan
\inst{1}\orcidlink{0009-0003-5667-926X} \and
Xi Wang\inst{1,5,8}\orcidlink{0000-0001-5442-1116} \and
Gerard Pons-Moll\inst{3,4}\orcidlink{0000-0001-5115-7794} \and
Matthias Nie\ss{}ner\inst{5}\orcidlink{0000-0001-6093-5199} \and
Marc Pollefeys\inst{1,7}\orcidlink{0000-0003-2448-2318} \and
Michael J. Black\inst{2}$^{\dagger}$\orcidlink{0000-0001-6077-4540} \and
Justus Thies\inst{6}$^{\dagger}$\orcidlink{0000-0002-0056-9825}}

\authorrunning{V.~Sklyarova et al.}

\institute{ETH Zürich, Switzerland \and 
Max Planck Institute for Intelligent Systems, Tübingen, Germany \and 
Max Planck Institute for Informatics, Saarland Informatics Campus, Germany \and 
Tübingen AI Center, University of Tübingen, Germany \and
Technical University of Munich, Germany
\and
Technical University of Darmstadt, Germany
\and Microsoft \and MCML
}

\renewcommand\twocolumn[1][]{#1}%
\maketitle
\begin{center}
    \centering
    \captionsetup{type=figure}
     \includegraphics[width=0.93\textwidth,trim=0 2.3cm 0 0,clip]{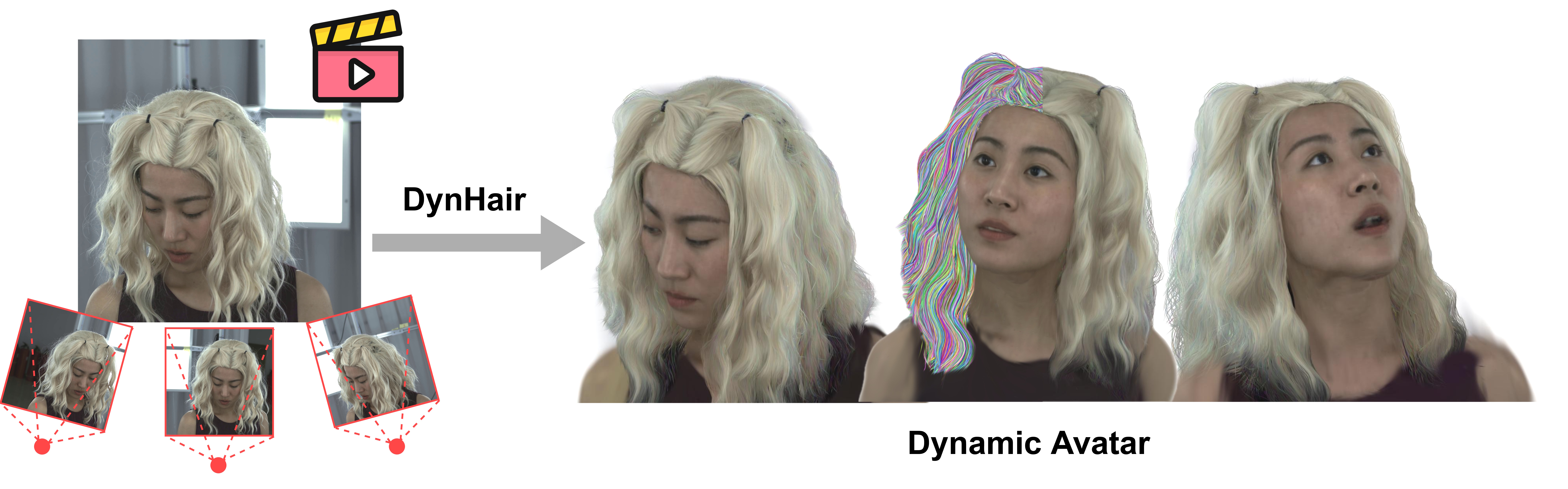}
    \caption{
		DynHair is a novel method for learning hair dynamics of 3D strands from data.
        It leverages multi-view video input to learn a dynamic hair strand deformation model based on the velocity and acceleration of the human head.
	}
    \label{fig:teaser}
\end{center}

\begin{abstract}
    We present DynHair, a novel method for tracking and modeling dynamic hair for human head avatars.
From video input, we reconstruct a dynamic head avatar with an explicit strand-based hair representation using structured 3D Gaussian Splatting.
In contrast to the face region of human head avatars, which can be modeled with 3D Gaussians that are attached or generated with respect to some expressive 3D head model, hair is particularly challenging as it exhibits dynamic motion effects.
Therefore, we present a novel method that models the dynamic deformations of the hair strands using a temporal network that is conditioned on angular velocity and acceleration of the head, as well as relative gravity.
Specifically, an LSTM encodes the motion history and modulates per-point strand features via FiLM conditioning which is then used by an MLP to produce physically plausible displacements to the canonical hairstyle.
We jointly optimize this motion and appearance representation of the hair, with a 3DGS-based representation of the face-region, via differentiable Gaussian splatting with photometric, geometric, and physics-based supervision.
As a result, we obtain hair tracking of the training video data and an animatable head avatar with controllable hair dynamics.
In our experiments, we demonstrate state-of-the-art performance in terms of hair dynamics, temporal consistency, and generalization across subjects. 
Our data and code are available at \url{https://dynhair.is.tue.mpg.de/}.

    \keywords{Digital Humans \and  Dynamic Hair Modeling \and  Hair Tracking}
\end{abstract}

\section{Introduction}
In the past few years, there has been tremendous progress in modeling photo-realistic 3D digital human appearance for applications in e-commerce, entertainment, and telepresence in AR/VR~\cite{zielonka2026star}.
Especially, the introduction of neural radiance fields~\cite{mildenhall2020nerf} and 3D Gaussian splatting~\cite{kerbl3Dgaussians} has led to a series of works ~\cite{hong2022headnerf,kabadayi24ganavatar, chen2024monogaussianavatar, wang2025gaussianhead, Teotia_audiodriven} focusing on visual quality~\cite{xu2023gaussianheadavatar,kirschstein2023nersemble, giebenhain2024npga, saito2024relightable}, runtime~\cite{zielonka2023instant,zielonka2024gem, SqueezeMe}, few-shot reconstruction~\cite{chu2024gagavatar, zielonka2025synshot, kirschstein2025avat3r}, and disentanglement~\cite{feng2023learning,wang2024mega}.
In addition to photorealism, realistic motion~\cite{thambiraja2025diface, aneja2024gaussianspeech, danecek2025supervising3dtalkinghead} is a critical aspect for immersive avatars. 
Avatars should behave like their physical counterparts. 
However, current avatar methods concentrate mostly on the face region, where they use 3D morphable model-based~\cite{li2017flame, paysan20093d} conditioning to handle expression-dependent deformations, ignoring dynamic hair motion.
The lack of dynamic hair breaks the sense of realism as the avatars do not follow physical principles.
When the head rotates to the side, the hair should follow the direction of gravity.
When the head shakes, the hair should lag behind, swing with inertia, and exhibit physically plausible follow‑through and settling behavior.

Such behavior can be modeled with physics-based simulation~\cite{maya, Blender}, requiring a strand-based reconstruction of the hair and appropriate physics parameters for mass, stiffness, damping, friction and more.
Recovering such a strand-based reconstruction is highly challenging as most of the hair strands are occluded and only the outer hair layer is visible.
Existing works that aim at reconstructing strand-based hair~\cite{sklyarova2025im2haircut, zakharov2024gh, Wu2024monohair, He2025perm, Radu2025difflocks, BenAyed2026Vid2Haircut} from images and video assume static hair geometry and exploit 
learned hairstyle priors that help reconstruct the inner structure of the hair.
They demonstrate animation with such reconstructed hairstyles, but the physics parameters for simulation are hand-tuned and not recovered from the observations.
Automatically finding realistic physics parameters for simulation is currently an unsolved problem, as it would require efficient differentiable hair simulation.
In contrast, we propose to learn hair dynamics directly from data with physics-based losses.

Specifically, we propose \textit{DynHair}, a method to learn dynamic strand-based hair directly from multi-view video.
It captures appearance through strand-aligned 3D Gaussians that follow the underlying hair geometry.
The canonical hairstyle is initialized from a strand-based hair prior~\cite{sklyarova2025im2haircut} and deformed at each timestep based on the head motion history.
The key component is the temporal motion model for the hair that is conditioned on gravity, and the velocity and acceleration of the head.
Specifically, we extract angular velocity, acceleration, and relative gravity from 3D head tracking and encode a sliding window of these to condition an LSTM to produce a temporal motion embedding.
Based on this motion embedding, we use FiLM~\cite{perez2018film} conditioning to modulate per-point strand features before a strand MLP predicts displacements of the 3D Gaussians.
A learnable root attenuation factor encodes the physical prior that strand roots anchored to the scalp move less than free-moving tips.
To train the method end-to-end, we combine photometric losses (RGB, SSIM, perceptual) with hair-geometric supervision (silhouette, orientation, penetration), physics-based regularization (elastic stretch preservation), and color regularization for spatially and temporally coherent hair appearance.
In addition to the hair, we jointly render and train the upper body region as an unstructured set of 3D Gaussians, whose positions and attributes are conditioned on facial expression and pose via deformation MLPs following \cite{xu2023gaussianheadavatar}.

\smallskip%
Our approach combines the rendering quality of Gaussian-based avatars with the geometric fidelity and physical plausibility of explicit strand representations.
The experiments demonstrate state-of-the-art results on dynamic head avatars, with improved hair dynamics, temporal consistency, and generalization across subjects and motion patterns compared to existing methods.
In summary, our main contributions are: 
\begin{itemize}
    \item We propose a dynamic hair motion model based on structured hair strands. An LSTM encoder processes head-local angular velocity, acceleration, and relative gravity as temporal conditioning signals. The resulting pose features modulate canonical hair point features via FiLM layers, which are then passed to an MLP to predict non-rigid hair motion.
    
    \item We propose an avatar reconstruction approach that disentangles hair and upper body regions, with region-dependent motion modeling and appearance representations learned from multi-view video.
\end{itemize}

\section{Related Work}

\noindent\textbf{Head avatars.}
Early head avatar approaches rely on explicit geometric representations, including meshes~\cite{weise2011realtime, blanz2003reanimating, ichim2015dynamic, bouazizrealtimeface, grassal2022neural} and volumetric primitives~\cite{lombardi2021mixture,lombardi2021voltemorph}. 
Subsequently, following NeRF~\cite{mildenhall2020nerf}, implicit representations~\cite{mildenhall2020nerf,mescheder2019occupancy,yariv2021volume} have become predominant~\cite{zheng2022imavatar,park2021nerfies,park2021hypernerf,gafni2021nerface,gao2022reconstructing, kirschstein2023nersemble,zielonka2023instant, nehvi2024volumetric, kabadayi24ganavatar}. Initial efforts focus on deformable NeRFs; for instance, Park \etal~\cite{park2021hypernerf} learn a hyperspace that maps observation space to a canonical space via a learnable latent code. 
Parametric face models such as FLAME~\cite{li2017flame} and BFM~\cite{paysan20093d} are widely used to guide the deformation from observation space to canonical space~\cite{gafni2021nerface,gao2022reconstructing,zielonka2023instant}. For example, Kirschstein \etal~\cite{kirschstein2023nersemble} introduce a deformation field and an ensemble of 3D multi-resolution hash encodings to construct high-fidelity human heads from multi-view videos. IMavatar~\cite{zheng2022imavatar} learns an expression-conditioned occupancy network from monocular video similar to SNARF~\cite{chen2021snarf}.

Although implicit representations achieve high realism, they require dense sampling in 3D space, which limits rendering efficiency. 
To overcome this, recent works use Gaussian Splatting~\cite{kerbl3Dgaussians} for head avatar modeling~\cite{xu2023gaussianheadavatar,qian2024gaussianavatars,shao2024splattingavatar,xiang2024flashavatar, giebenhain2024npga, kirschstein2025avat3r,chen2024monogaussianavatar, wang2024mega, wang2025gaussianhead}, offering real time rendering while maintaining visual fidelity.
Gaussian Head Avatar (GHA)~\cite{xu2023gaussianheadavatar} conditions Gaussians on BFM parameters and incorporates a super resolution module for high quality rendering. 
GaussianAvatars~\cite{qian2024gaussianavatars} attaches Gaussians to a rigged FLAME mesh to enable controllable animation. SplattingAvatar~\cite{shao2024splattingavatar} embeds Gaussians on a triangular mesh using barycentric coordinates. Similarly, FlashAvatar~\cite{xiang2024flashavatar} embeds Gaussians in the UV space of a FLAME mesh with a learnable offset for fine details. 
GaussianHead~\cite{wang2025gaussianhead} uses a parametric face mesh to describe large scale motion and stores view-dependent appearance in tri-planes. 
Unlike our approach, all the above methods treat hair implicitly as part of the head representation, limiting strand-level detail and hair motion.
We disentangle the hair and the face, introducing a new hair model, and represent the head following GHA~\cite{xu2023gaussianheadavatar}.

\noindent\textbf{Hair Animation.}
Hair motion plays a crucial role in character animation, games, and virtual avatars, yet remains challenging due to the large number of strands, complex physical behavior, and frequent self- and body-collisions. 
Methods that model hair dynamics can be broadly categorized into physics-based simulators~\cite{Bertails2006super,Herrera2024augmented,daviet2023interactive,Huang2023towards}, approaches that approximate simulator behavior with learned models~\cite{Zhou2023groomgen, wang2025dgh,Stuyck2024quaffure}, and methods that learn dynamics from image observations~\cite{Yang2019dynamic,Wang2022hvh,Wang2023neuwigs,Liao2025hhavatar}. 

\noindent\textit{Physics-based simulators} generate hair dynamics using numerical solvers and are widely integrated in production~\cite{maya,unrealengine,Blender,He2024digitalsalon}. While stable and interpretable, they are computationally expensive and require manual parameter tuning for each hairstyle type.

\noindent\textit{Learned approximations of simulation.} Recent methods, such as GroomGen~\cite{Zhou2023groomgen} and DGH~\cite{wang2025dgh}, approximate the simulator's behavior by training a neural network on 3D data generated in Houdini.
Quaffure~\cite{Stuyck2024quaffure} extends this direction with self-supervised learning of hair dynamics conditioned on body pose using physics-based constraints. However, none of these methods are personalized, nor do they model full avatar appearance.

\noindent\textit{Capture-based methods} learn dynamics directly from an observed video. Yang et al.~\cite{Yang2019dynamic} reconstruct dynamic hair from a monocular video using separate networks for spatial occupancy and temporal motion.
HVH~\cite{Wang2022hvh} represents hair volumetrically %
and learns dynamics via guide-strand tracking with RGB and optical flow supervision.
NeuWigs~\cite{Wang2023neuwigs} learns a compressed appearance representation followed by a temporal transfer module for latent-space dynamics.
HHAvatar~\cite{Liao2025hhavatar} models hair with unstructured Gaussians and predicts deformations via an MLP conditioned on past hair states and head poses. 
PhysHead~\cite{kabadayi2026physhead} first reconstructs a static strand-based hairstyle using Neural Haircut~\cite{Sklyarova2023NeuralHP}, then animates the resulting Gaussians with dynamics from the physics-based simulator Maya. While PhysHead optimizes hair appearance from a static time step, we optimize from a dynamic sequence.
Concurrent work HADES~\cite{Liao_2025_HADES} also uses strand-based geometry and learns dynamics conditioned on head pose for full-body avatars, but conditions on a history of past hair states and models only a sparse set of strands, interpolating the rest via nearest neighbors.

In contrast to these capture-based methods, we model hair motion in a structured manner using a strand-based representation. We predict deformations for all strands relative to canonical hair, rather than from a history of past states, and optimize appearance directly from a dynamic sequence rather than a static one. We further incorporate acceleration and gravity as explicit inputs and introduce a physics-based elastic term to improve the realism of the resulting motion.

\begin{figure*}[t]
    \centering
    \includegraphics[width=\textwidth]{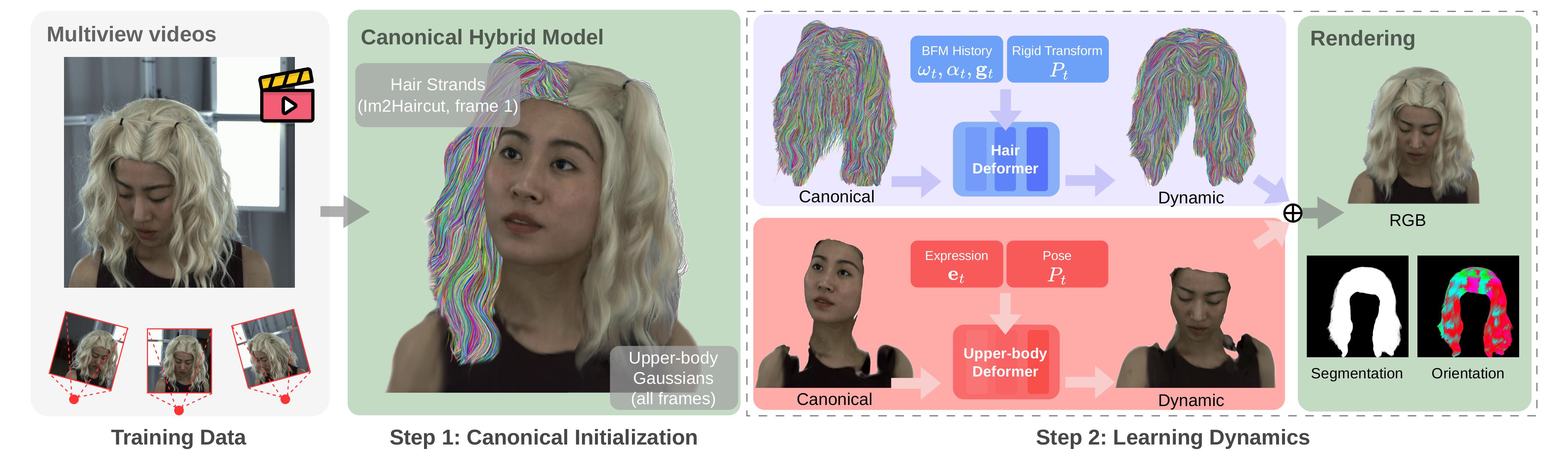}
    \caption{
        \textbf{Method overview.} 
        Given multi-view video input, we first initialize a canonical hybrid model for the upper body and hair.
        Hair strands are initialized using the Im2Haircut~\cite{sklyarova2025im2haircut} prior model, and the upper body is represented with unstructured 3D Gaussians~\cite{xu2023gaussianheadavatar}.
        We then jointly optimize hair and head deformations with two separate motion models.
        The hair deformation is conditioned on BFM motion history with an additional rigid transformation, while the head deformation is conditioned on facial expressions and pose parameters.
        Finally, the hair and upper body Gaussians are composited and rendered via differentiable splatting to produce the output images.
    }
    \label{fig:pipeline}
\end{figure*}

\section{Method}
Our method models a dynamic head avatar with explicit strand-based hair from multi-view videos. 
As shown in \Cref{fig:pipeline}, our proposed method disentangles hair and the upper body.
Both regions, and especially their deformations, are handled differently.
Following~\cite{xu2023gaussianheadavatar}, the upper body is represented with an unstructured set of 3D Gaussians whose positions and attributes are conditioned on facial expression and pose.
In contrast to \cite{xu2023gaussianheadavatar}, which does not disentangle body and hair, we model the hairstyle separately as a set of explicit polyline strands with strand-aligned Gaussian primitives and a learned dynamic deformation module that is conditioned on the head motion history and gravity. 
At each timestep, we predict deformations of both the upper body and the hair, concatenate all their Gaussians, and render them jointly via differentiable Gaussian Splatting~\cite{kerbl3Dgaussians} to obtain RGB color, segmentation labels, and strand orientation maps enabling joint supervision through photometric losses, hair-geometric constraints, and physics-based regularization.
In the following, we introduce our novel hair modeling, while the upper body model is detailed in the \suppl.

\subsection{Modeling Dynamic Hair}
Our dynamic hair modeling consists of two stages: static hairstyle reconstruction at the initial timestep, followed by learning a head motion-conditioned deformation network for dynamic animation of hair.

~

\noindent\textbf{Canonical representation.}
We model the hairstyle $\mathcal{H}$ as a set of $N$ polylines originating from a scalp mesh: $\mathcal{H} = \{S_i\}_{i=1}^{N}$, where each strand $S_i = \{p_{ij}\}_{j=1}^{L}$ consists of $L$ points.  To obtain the canonical hairstyle, we leverage Im2Haircut~\cite{sklyarova2025im2haircut}, a PCA-based hair prior with learned coarse-to-fine encoders.
Given multi-view images at the initial timestep ($t{=}0$), the encoder predicts PCA coefficients that are decoded into strand geometry through a learned shape basis. The resulting strands are upsampled to the target resolution, filtered using a scalp silhouette mask to remove spurious strands, and then resampled to $L=40$ points per strand.
We jointly optimize the strand geometry using the Im2Haircut reconstruction losses and additional regularization for geometric consistency:
\begin{equation}
    \mathcal{L}_\text{static} = \mathcal{L}_\text{im2haircut} + \lambda_\text{pca}\mathcal{L}_\text{pca} + \lambda_\text{length}\mathcal{L}_\text{length} + \lambda_\text{smooth}\mathcal{L}_\text{smooth} + \lambda_\text{consist}\mathcal{L}_\text{consist} ,
\end{equation}
where $\mathcal{L}_\text{pca}$ regularizes PCA coefficients, $\mathcal{L}_\text{length}$ and $\mathcal{L}_\text{smooth}$ encourage nearby strands to have similar lengths and shapes (measured via root-based KNN), and $\mathcal{L}_\text{consist}$ enforces that the PCA distribution of back-view strands matches that of front-view strands, improving geometry in occluded regions. This yields the canonical hairstyle $\mathcal{H}_0$ used as input for the dynamic stage.

~

\noindent\textbf{Motion conditioning.}
Hair dynamics result from both rigid head motion and non-rigid deformations caused by inertia and gravity. 
To model these dynamics, we extract the global pose $P_t = \{R_t, T_t, S_t\}$ from the parametric face model~\cite{paysan20093d} at each timestep $t$ and compute the \emph{angular velocity, acceleration}, and \emph{relative gravity direction}.
The \emph{angular velocity and acceleration} of the head are computed using central differences. 
At each timestep $t$, they are given by:
\begin{equation}
    \boldsymbol{\omega}^\text{world}_t = \frac{\mathrm{Log}\!\left(R_{t+1}\, R_{t-1}^\top\right)}{2\,\Delta t},
    ~~~~~
    \boldsymbol{\alpha}^\text{world}_t = \frac{\boldsymbol{\omega}^\text{world}_{t+1} - \boldsymbol{\omega}^\text{world}_{t-1}}{2\,\Delta t},
\end{equation}
where $\mathrm{Log}(\cdot)$ converts a rotation from SO(3) to $\mathbb{R}^3$.
Both signals are then rotated into the head-local frame to obtain pose-invariant conditioning: $\boldsymbol{\omega}_t = R_t^\top \boldsymbol{\omega}^\text{world}_t$ and $\boldsymbol{\alpha}_t = R_t^\top \boldsymbol{\alpha}^\text{world}_t$. 
The \emph{relative gravity direction}, defined as the gravity vector in the head-local frame, is obtained by rotating the canonical gravity vector:
\begin{equation}
    \mathbf{g}_t = \mathrm{normalize}\!\left(R_t^\top\, [0,\, {-1},\, 0]^\top\right).
\end{equation}
For training and inference stability, we normalize $\boldsymbol{\omega}_\tau$ and $\boldsymbol{\alpha}_\tau$ per-axis by precomputed maximum absolute values from the training set and clamped to $[-1, 1]$. 

~

\noindent\textbf{Hair deformation network.}
We train a hair deformer network $\mathcal{D}_\text{hair}$ that predicts non-rigid per-point displacements for the canonical hairstyle, conditioned on the head motion history of the past $T$ frames. 
Specifically, for each frame $\tau$ in the sliding window $\{t{-}T, \ldots, t\}$, we form the conditioning vector $\mathbf{c}_\tau = [\boldsymbol{\omega}_\tau, \boldsymbol{\alpha}_\tau, \mathbf{g}_\tau] \in \mathbb{R}^{9}$.
Each $\mathbf{c}_\tau$ is mapped through positional encoding $\gamma(\mathbf{c}_\tau)$ and then processed by an LSTM with learnable initial hidden states $(h_0, c_0)$ to produce a temporal motion embedding $\mathbf{z}_t$:
\begin{align}
\mathbf{z}_t &= \mathrm{LSTM}\!\left(\{\gamma(\mathbf{c}_\tau)\}_{\tau=t-T}^{t};\; h_0, c_0\right).
\end{align}
The motion embedding $\mathbf{z}_t$ modulates per-point strand features via Feature-wise Linear Modulation (FiLM)~\cite{perez2018film}. A linear layer maps $\mathbf{z}_t$ to scale and shift parameters $(\boldsymbol{\gamma}_\text{film}, \boldsymbol{\beta}_\text{film})$, which modulate the positional encoding of strand point $p_{ij}$:
\begin{align}
(\boldsymbol{\gamma}_\text{film}, \boldsymbol{\beta}_\text{film}) &= \mathrm{Linear}(\mathbf{z}_t), &
\mathrm{FiLM(p_{ij})} &= \gamma(p_{ij}) \odot (1 + \boldsymbol{\gamma}_\text{film}) + \boldsymbol{\beta}_\text{film} .
\end{align}
The modulated features are concatenated with a strand position embedding (encoding the normalized position $j/L$ along the strand) and passed through a strand-wise MLP to predict per-point displacements:
\begin{equation}
    \Delta p_{ij} = \mathrm{MLP}\Big(\mathrm{FiLM(p_{ij})}, \gamma(j/L)\Big) \cdot \rho_j ,
\end{equation}
where $\rho_j$ is a learnable root attenuation factor initialized as a linear ramp from small values near the root to 1.0 at the tip. This encodes the physical prior that strand roots anchored to the scalp exhibit less displacement than free-moving tips. The final MLP layer is initialized near zero to ensure small initial deformations and stable early training. The deformed hairstyle is:
\begin{equation}
    \mathcal{H}_t = \mathcal{H}_0 + \Delta\mathcal{H}_t,
\end{equation}
which is rigidly transformed to world coordinates using the head's global pose.

~

\noindent\textbf{Strand-aligned Gaussian rendering.}
Following~\cite{zakharov2024gh, Luo2024gaussianhair, Zhou2024groomcap}, we constrain the 3D Gaussian primitives to lie on the strand polylines ${S_i}$. The mean of each Gaussian is placed at the midpoint of a line segment, its primary scale axis is aligned with the segment direction (with magnitude proportional to the segment length), and the two cross-sectional scale axes are set to a fixed strand width. The Gaussian rotation quaternion is computed via parallel transport from the reference axis to the segment direction. We taper the Gaussians near strand roots to reduce visual artifacts at the scalp boundary.

\subsection{Joint Training and Losses}
We concatenate the 3D Gaussians for the hair and upper body and jointly render them via differentiable Gaussian Splatting~\cite{kerbl3Dgaussians} to obtain RGB images, segmentation masks for hair and body, and orientation maps for hair. 
Training is performed in two stages:
(i) the \emph{static initialization} stage, where we optimize for the canonical hairstyle $\mathcal{H}_0$ using the first timestep across all views; 
and (ii) the \emph{dynamic training} stage, where we jointly train the hair deformation network $\mathcal{D}_\text{hair}$, $\mathcal{H}_0$ with small learning rate, head deformation networks, and hair appearance.  
The objective combines photometric, hair-geometric, physics-based, and color-regularization losses:
\begin{equation}
     \mathcal{L}_\text{dynamic} = \mathcal{L}_\text{photo} + \mathcal{L}_\text{hair} + \mathcal{L}_\text{color\_reg} .
\end{equation}

\noindent\textbf{Photometric losses.}
We supervise the rendered images using a combination of pixel-wise, structural similarity, and perceptual losses:
\begin{equation}
    \mathcal{L}_\text{photo} = \lambda_\text{rgb}\mathcal{L}_\text{rgb} + \lambda_\text{ssim}\mathcal{L}_\text{ssim} + \lambda_\text{vgg}\mathcal{L}_\text{vgg},
\end{equation}
where $\mathcal{L}_\text{rgb}$ is the L1 distance between the rendered and ground-truth images.
The structural dissimilarity loss $\mathcal{L}_\text{ssim} = 1 - \mathrm{SSIM}(\hat{I}, I)$ penalizes deviations in local image structure using a Gaussian-windowed SSIM computation. 
The perceptual loss $\mathcal{L}_\text{vgg}$ is the LPIPS distance~\cite{zhang2018unreasonable} computed with a pre-trained VGG backbone, encouraging realistic texture and high-frequency detail.

\smallskip
\noindent\textbf{Geometric losses.}
To supervise geometry, we introduce losses on the silhouette, strand orientation, head-hair penetration and inextensibility constraints:
\begin{equation}
    \mathcal{L}_\text{hair} = \lambda_\text{seg}\mathcal{L}_\text{seg} + \lambda_\text{orient}\mathcal{L}_\text{orient} + \lambda_\text{penetr}\mathcal{L}_\text{penetr} + \lambda_\text{elastic}\mathcal{L}_\text{elastic} .
\end{equation}
The \emph{segmentation loss} $\mathcal{L}_\text{seg}$ encourages the rendered hair silhouette to match the ground-truth hair mask $\hat{M}_h$. We use an asymmetric recall formulation that penalizes missing ground-truth coverage more strongly than over-coverage:
\begin{equation}
    \mathcal{L}_\text{seg} = \frac{1}{|\Omega|}\sum_\Omega \max\!\left(\hat{M}_h - M_h,\, 0\right) + \frac{0.5}{|\Omega|}\sum_\Omega \max\!\left(M_h - \hat{M}_h,\, 0\right),
\end{equation}
where $M_h$ is the predicted hair silhouette and $\Omega$ denotes the set of visible pixels. An exponential decay schedule is applied after a warm-up period to avoid over-constraining the silhouette.
The \emph{orientation loss} $\mathcal{L}_\text{orient}$ matches predicted 2D strand orientations with ground-truth orientation maps. Since undirected orientations have $180^\circ$ ambiguity, we use a wrap-aware formulation:
\begin{equation}
    \mathcal{L}_\text{orient} = \frac{\pi}{|\Omega_h|}\sum_{\Omega_h}\min\!\left(|\theta_\text{pred} - \theta_\text{gt}|,\; |\theta_\text{pred} - \theta_\text{gt} - 1|,\; |\theta_\text{pred} - \theta_\text{gt} + 1|\right),
\end{equation}
where $\Omega_h$ is the intersection of predicted and ground-truth hair masks.
The \emph{penetration loss} $\mathcal{L}_\text{penetr}$ prevents hair strands from penetrating the head mesh. We randomly sample up to $10{,}000$ strand points and use signed queries to identify points inside the head mesh $\mathcal{M}$. For inside points, we penalize the squared distance to the nearest mesh triangle:
\begin{equation}
    \mathcal{L}_\text{penetr} = \frac{1}{|\mathcal{P}_\text{in}|}\sum_{p \in \mathcal{P}_\text{in}} d^2(p,\, \mathcal{M}), \quad \mathcal{P}_\text{in} = \left\{p_{ij} : \mathrm{inside}(p_{ij},\, \mathcal{M}) \right\},
\end{equation}
where $d^2(p, \mathcal{M})$ denotes the squared point-to-mesh distance.
The \emph{elastic regularization term}
penalizes stretching and compression of hair segments during deformation. For consecutive strand points $p_j$ and $p_{j+1}$, let $d^\text{rest} = p_{j+1}^0 - p_j^0$ and $d^\text{posed} = p_{j+1}^t - p_j^t$ denote the segment vectors in the rest and posed configurations, respectively. We apply this loss only after static initialization:%
\begin{equation}
    \mathcal{L}_\text{elastic} = \frac{1}{N L}\sum_{i=1}^{N}\sum_{j=1}^{L-1}\left(\|d_{ij}^\text{posed}\| - \|d_{ij}^\text{rest}\|\right)^2 .
\end{equation}

\noindent\textbf{Hair color regularization.}
We introduce three complementary color regularization terms to ensure spatially and temporally coherent hair appearance:
\begin{equation}
    \mathcal{L}_\text{color\_reg} = \lambda_\text{cg}\mathcal{L}_\text{cg} + \lambda_\text{cs}\mathcal{L}_\text{cs} + \lambda_\text{cc}\mathcal{L}_\text{cc} .
\end{equation}
The \emph{color gradient loss} $\mathcal{L}_\text{cg}$ penalizes abrupt color changes between consecutive points along each strand: $\mathcal{L}_\text{cg} = \frac{1}{NL}\sum_{i,j}\|c_{i,j+1} - c_{i,j}\|^2$.
The \emph{spatial color smoothness loss} $\mathcal{L}_\text{cs}$ encourages spatially nearby Gaussians (across different strands) to share similar colors, computed via k-nearest-neighbor ($k{=}8$) matching: $\mathcal{L}_\text{cs} = \frac{1}{|\mathcal{N}|}\sum_{(a,b) \in \mathcal{N}}\|c_a - c_b\|^2$.
The \emph{color consistency loss} $\mathcal{L}_\text{cc}$ encourages uniform color along each strand by penalizing the deviation from the per-strand mean color: $\mathcal{L}_\text{cc} = \frac{1}{NL}\sum_{i,j}\|c_{i,j} - \bar{c}_i\|^2$. See \suppl for loss weights.

\section{Experiments}

\noindent\textbf{Dataset details.} 
We train DynHair on two multiview datasets: three scenes from the HHAvatar dataset~\cite{Liao2025hhavatar} and two from a newly captured dataset. %
HHAvatar has 4 synchronized cameras, about 3 minutes per scene at 2048 × 2048 resolution, with diverse hair motion and facial expressions.
Our new dataset captures actors with 15 synchronized cameras following a setup similar to~\cite{kirschstein2023nersemble}.
It includes a predefined set of diverse hair motions, spanning slow to rapid accelerations, nodding, tilting, rotations, lateral moves, sudden stops, with varied facial expressions. 
All participants performed the same 22 actions, 10 targeting hair dynamics.
Sequences are at $4k$ and $72$ FPS. We provide an additional 360° coverage with and without hairnet for static geometry estimation.

\smallskip
\noindent\textbf{Training details.} 
We train DynHair in three stages. In Stage~1, we train a mesh-based head model to obtain canonical head geometry. In Stage~2, we first reconstruct the canonical hairstyle using Im2Haircut~\cite{sklyarova2025im2haircut} from all views and then 
we train the hair deformation and optimize the head and hair Gaussian primitives jointly. We train our full model for 320,000 iterations on a single A100, increasing the VGG perceptual loss weight for the last 80,000 iterations to enhance fine-grained appearance details. In all experiments, we use a resolution $1024 \times{1024}$, window history of 5 frames, 40 points per strand, and around 11,000 strands depending on hairstyle. See details in \suppl.

\subsection{Comparison and Results} 
\noindent\textbf{Baselines.} 
To our knowledge, there is no publicly available method that learns hair dynamics from data for direct comparison.
Therefore, we evaluate our approach against state-of-the-art multi-view dynamic head avatar methods, Gaussian Head Avatar~\cite{xu2023gaussianheadavatar} (GHA) and GaussianAvatars (GA)~\cite{qian2024gaussianavatars} as well as a physics simulator (Maya ~\cite{maya}) that computes hair dynamics via numerical integration of Newtonian forces. Since it outputs geometry only, we apply our estimated colors and drive Gaussians using the simulated motion.

\begin{figure*}[t]
    \centering
    \includegraphics[width=0.98\textwidth]{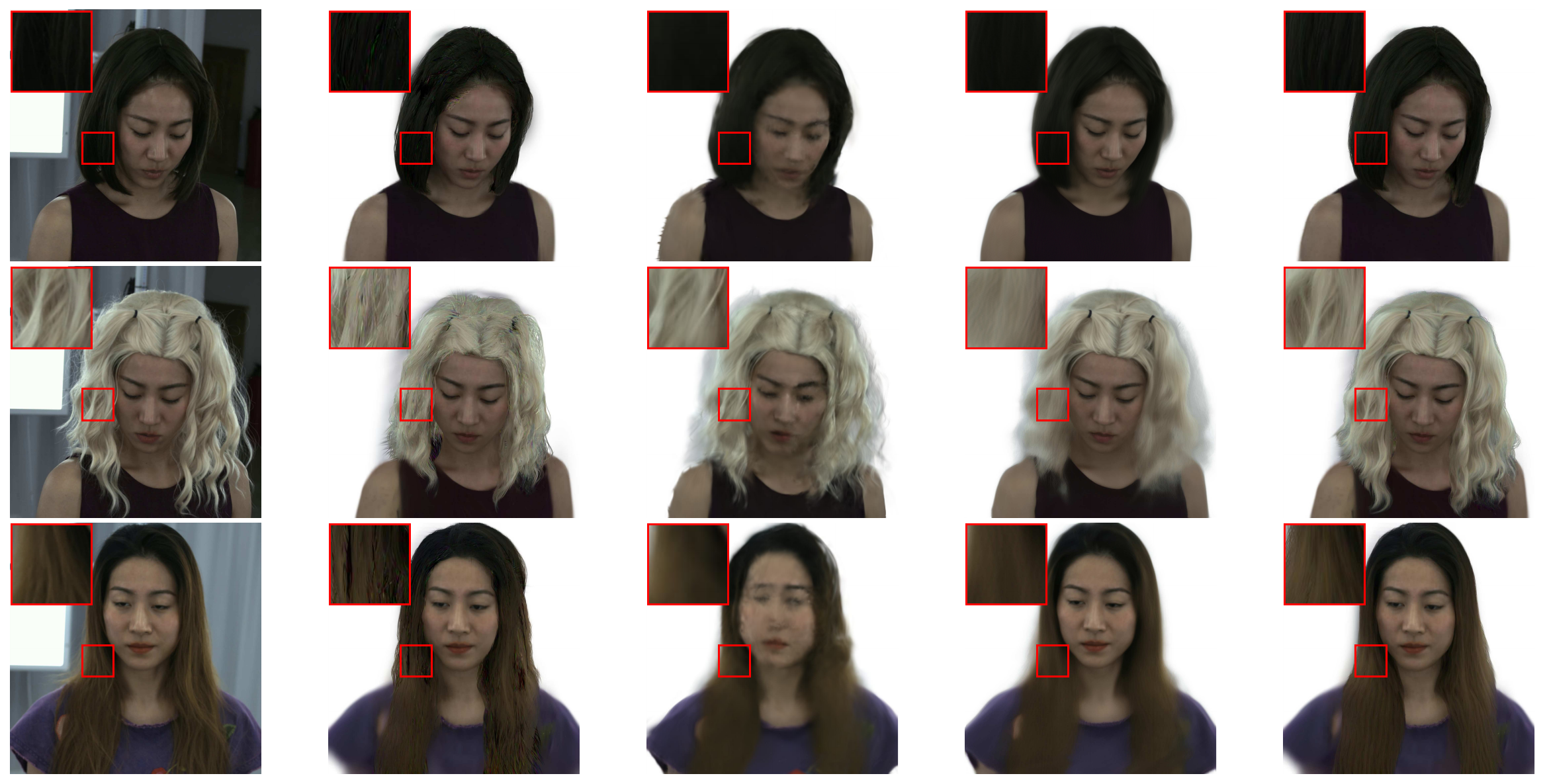}\\[2pt]
    \makebox[0.19\textwidth]{\footnotesize Input image}%
    \makebox[0.19\textwidth]{\footnotesize Simulator}%
    \makebox[0.19\textwidth]{\footnotesize GA~\cite{qian2024gaussianavatars}}%
    \makebox[0.19\textwidth]{\footnotesize GHA~\cite{xu2023gaussianheadavatar}}%
    \makebox[0.19\textwidth]{\footnotesize Ours}
    
    \caption{
        \textbf{Self-reenactment comparison.} Each row shows a different test frame or subject. Columns from left to right: ground-truth image, physics-based simulator from Maya~\cite{maya},  GaussianAvatars~\cite{qian2024gaussianavatars}, Gaussian Head Avatar~\cite{xu2023gaussianheadavatar}, and our method. Our approach produces sharper strand-level detail and more temporally coherent hair motion. 
    }
\label{fig:self_reenact}
\end{figure*}

\smallskip
\noindent\textbf{Qualitative evaluation.} 
We compare with the baselines 
 under both self-reenactment and cross-reenactment settings by evaluating appearance fidelity and motion realism in each scenario. 
Figure~\ref{fig:self_reenact} shows self-reenactment results on held-out test sequences. Our method produces sharper hair details and more temporally coherent strand motion than baselines, which produce over-smoothed and flickering hair regions. The physics-based simulator deviates from the true hair motion due to manually obtained parameters and the bounded accuracy of the physical model. Moreover, directly attaching learned appearance to simulator-driven strand positions results in degraded rendering quality (see Figure~\ref{fig:self_reenact}, middle row). This further motivates our joint learning approach.

\begin{figure*}[t]
    \centering
    \includegraphics[width=0.9\textwidth]{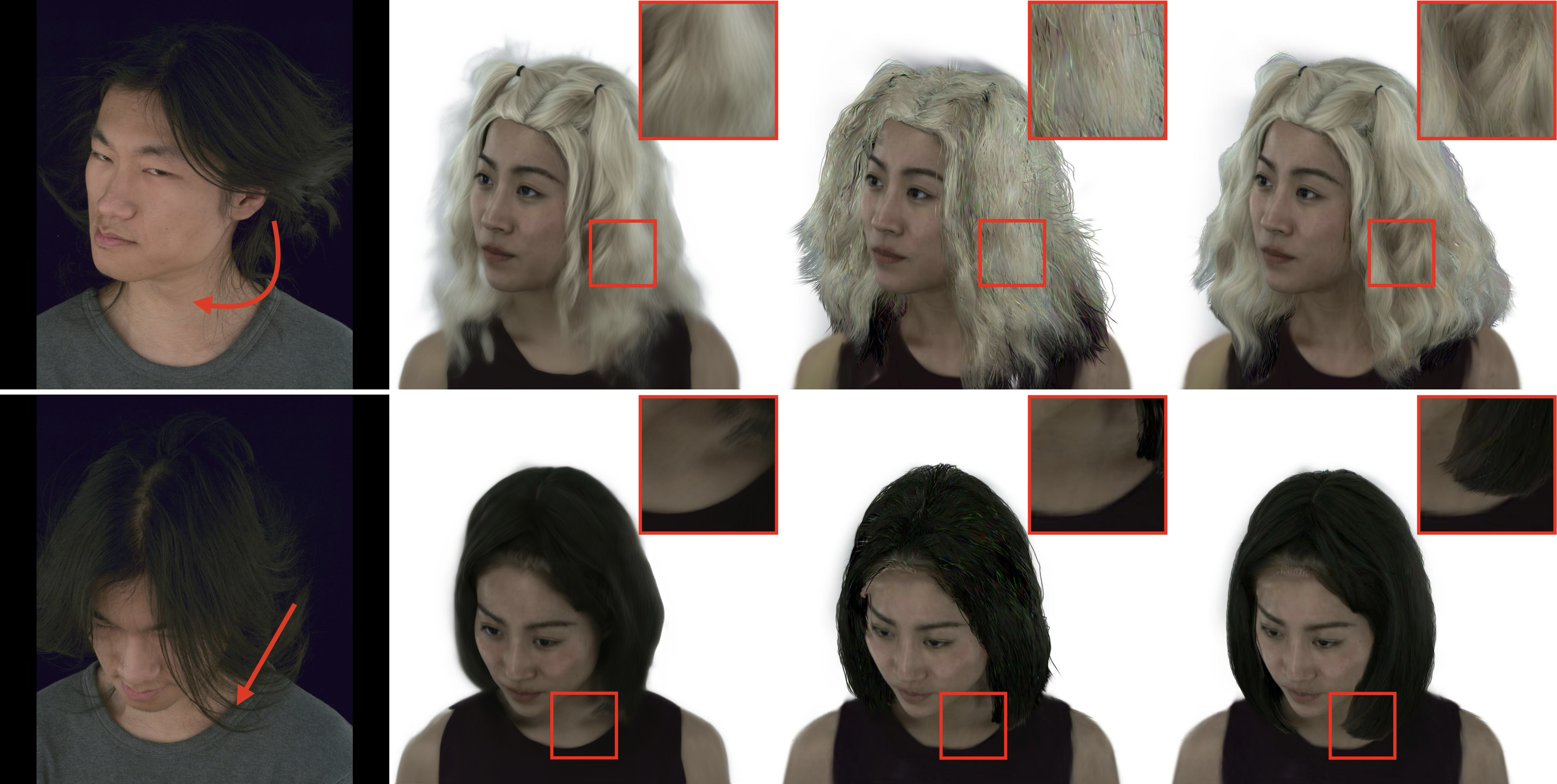}\\[2pt]
    \makebox[0.24\textwidth]{\footnotesize Reference}%
    \makebox[0.24\textwidth]{\footnotesize GHA~\cite{xu2023gaussianheadavatar}}%
    \makebox[0.24\textwidth]{Simulator}%
    \makebox[0.24\textwidth]{\footnotesize Ours}
    
    \caption{
        \textbf{Cross-reenactment comparison.} The 1st row shows left-right head rotation, while the second row corresponds to nodding. Our hairstyle better follows physics constraints and gravity. The image in the first column serves as a reference for human motion. While the hairstyle should follow general dynamics, it is not an exact match.}
    \label{fig:cross_reenact}
\end{figure*}

Figure~\ref{fig:cross_reenact} presents cross-reenactment results, where the driving signal comes from a different subject. 
Our explicit strand-based representation, conditioned on relative motion, produces geometry-consistent hair animation for novel driving sequences, while baselines struggle to preserve identity and realistic dynamics.

\begin{table*}[t]
\centering
\caption{\textbf{Quantitative comparison.} Self-reenactment results averaged across test sets of 3 subjects. As pixel-wise alignment metrics are insufficient to fully capture hair motion quality, we additionally report $tLPIPS_{\text{ex}}$ as a dedicated temporal measure. tIoU$_{\text{hair}}^{gt}$ is equal to 0.946, and $tLPIPS_{\text{ex}}$ should be close to 0.}

\label{tab:quantitative}
\setlength{\tabcolsep}{3pt}
\renewcommand{\arraystretch}{1.15}
\resizebox{\textwidth}{!}{%
\footnotesize
\begin{tabular}{l|cccc|cccccc|c|cc}
& \multicolumn{4}{c|}{Full Image}
& \multicolumn{6}{c|}{Hair Region}
& \multicolumn{1}{c|}{Temporal}
& \multicolumn{2}{c}{Motion ($\times 10^{-3}$)} \\
Method
& PSNR$\uparrow$ & SSIM$\uparrow$ & LPIPS$\downarrow$ & FID$\downarrow$
& PSNR$\uparrow$ & SSIM$\uparrow$ & LPIPS$\downarrow$ & IoU$_{\text{hair}}\uparrow$
& tIoU$_{\text{hair}}\uparrow$ & FID$\downarrow$
& tLPIPS$_{\text{ex}}$
& vel.$\uparrow$ & accel.$\downarrow$ \\
\hline
GA~\cite{qian2024gaussianavatars} & 20.17 & 0.7787 & 0.2536 & 45.73  & 19.94 & 0.7335 & 0.1362 & -- & -- & 126.93 & -0.0230 & 0.40 & 0.08   \\
GHA~\cite{xu2023gaussianheadavatar} & 22.33 & 0.7925 & 0.2056 &36.25 & 22.15 & 0.7446 & 0.1086 & -- & -- & 81.07 & -0.0127 & 2.17 & 0.24   \\
Maya* & 19.39 & 0.6610 & 0.2465 &62.64 & 17.89 & 0.5182 & 0.1484 & 0.776 & 0.925 & 121.97  & 0.0183 & 2.38 & 0.11   \\
Ours & 21.60 & 0.7638 & 0.2010 &30.06 & 21.01 & 0.6885 & 0.1101 & 0.878 & 0.936 & 37.51 & 0.0045 & 2.41 & 0.19   \\
\hline
\hline
\end{tabular}%
}
\end{table*}

\smallskip
\noindent\textbf{Quantitative evaluation.} 
For quantitative comparison, we report pixel-aligned metrics (PSNR, SSIM) alongside perceptual LPIPS~\cite{zhang2018unreasonable} and measure distribution realism (FID)~\cite{heusel2018ganstrainedtimescaleupdate}. We evaluate on both full images and the hair region, computed on a test set of three subjects. 
Our method achieves slightly lower PSNR and SSIM compared to GHA~\cite{xu2023gaussianheadavatar}. This behavior is expected: enforcing strand-level geometric consistency prevents the model from arbitrarily smoothing high-frequency structures. 
While this introduces small pixel-wise deviations that negatively affect reconstruction metrics, it better preserves fine strand-level details and overall geometric fidelity. The marginally higher hair LPIPS similarly reflects a fundamental metric mismatch, as LPIPS penalizes any deviation from a single GT strand configuration regardless of physical plausibility, and thus favors appearance memorization over dynamic correctness. As a method that explicitly models hair dynamics, ours generates temporally coherent, physically plausible motion rather than reproducing one specific GT realization. We show superior performance in terms of the FID metric, which measures perceptual realism and better reflects the goals of hair dynamics learning than pixel-aligned reconstruction scores.
To measure temporal flickering, we compute $\mathrm{tLPIPS}_{\text{ex}} =$  
\begin{equation}
\frac{1}{T-1}\sum_{t=1}^{T-1} \Big[
\,\mathrm{LPIPS}\!\left(
I_t \odot \hat{M}_t,\,
I_{t+1} \odot \hat{M}_t
\right) \nonumber
-\mathrm{LPIPS}\!\left(
\hat{I}_t \odot \hat{M}_t,\,
\hat{I}_{t+1} \odot \hat{M}_t
\right)
\Big],
\end{equation}
where $\hat{M}_t = \mathrm{clamp} \ \!(\hat{H}_t + \hat{H}_{t+1},\, 0,\, 1)$. 
Here, $I_t$ and $\hat{I}_t$ denote the predicted and ground-truth images at frame $t$, respectively, and $\hat{H}_t$ is the ground-truth hair segmentation mask at frame $t$. LPIPS between consecutive frames measures perceptual temporal variation. Subtracting the corresponding ground-truth temporal LPIPS isolates excess perceptual change introduced by the model, yielding a measure of temporal fidelity beyond natural motion. 
Values near zero indicate alignment with the ground truth, positive values indicate excess variation, and negative values indicate over-smoothing.

Temporal evaluation highlights the benefit of explicit strand-based modeling. Baselines produce negative $\mathrm{tLPIPS}_{\text{ex}}$, indicating over-smoothed, stiff motion, while our method better matches ground-truth hair dynamics.
These results demonstrate a principled trade-off: unconstrained Gaussians optimize pixel alignment at the cost of sharp high-frequency detail and realistic temporal behavior, while our structured representation prioritizes geometric and temporal consistency, producing sharper and more dynamically faithful hair reconstructions while maintaining competitive global image quality.

We also compute the non-rigid hair velocity and acceleration by unposing the head to a fixed canonical pose. Then, we define velocity as the mean nearest-neighbor displacement 
between consecutive hair point clouds in this head-local frame, and acceleration as its finite difference. For our method and simulator, we use the structured strand vertices; for baselines with unstructured Gaussians, we select those whose projections fall within the hair silhouette mask at the first frame and use the same fixed indices across all subsequent frames. Both representations are subsampled to the same point count for fair comparison.  Our method produces higher hair velocity, indicating more expressive hair motion  (see~\cref{tab:quantitative}), compared to baselines with unstructured representation. The Maya simulator achieves lower acceleration, demonstrating more temporally coherent hair dynamics; however, there is increased error in IoU$_{\text{hair}}$
and tIoU$_{\text{hair}}$.

\begin{figure*}[t]
    \centering
    \vspace{-0.25cm}
    \includegraphics[width=0.96\textwidth]{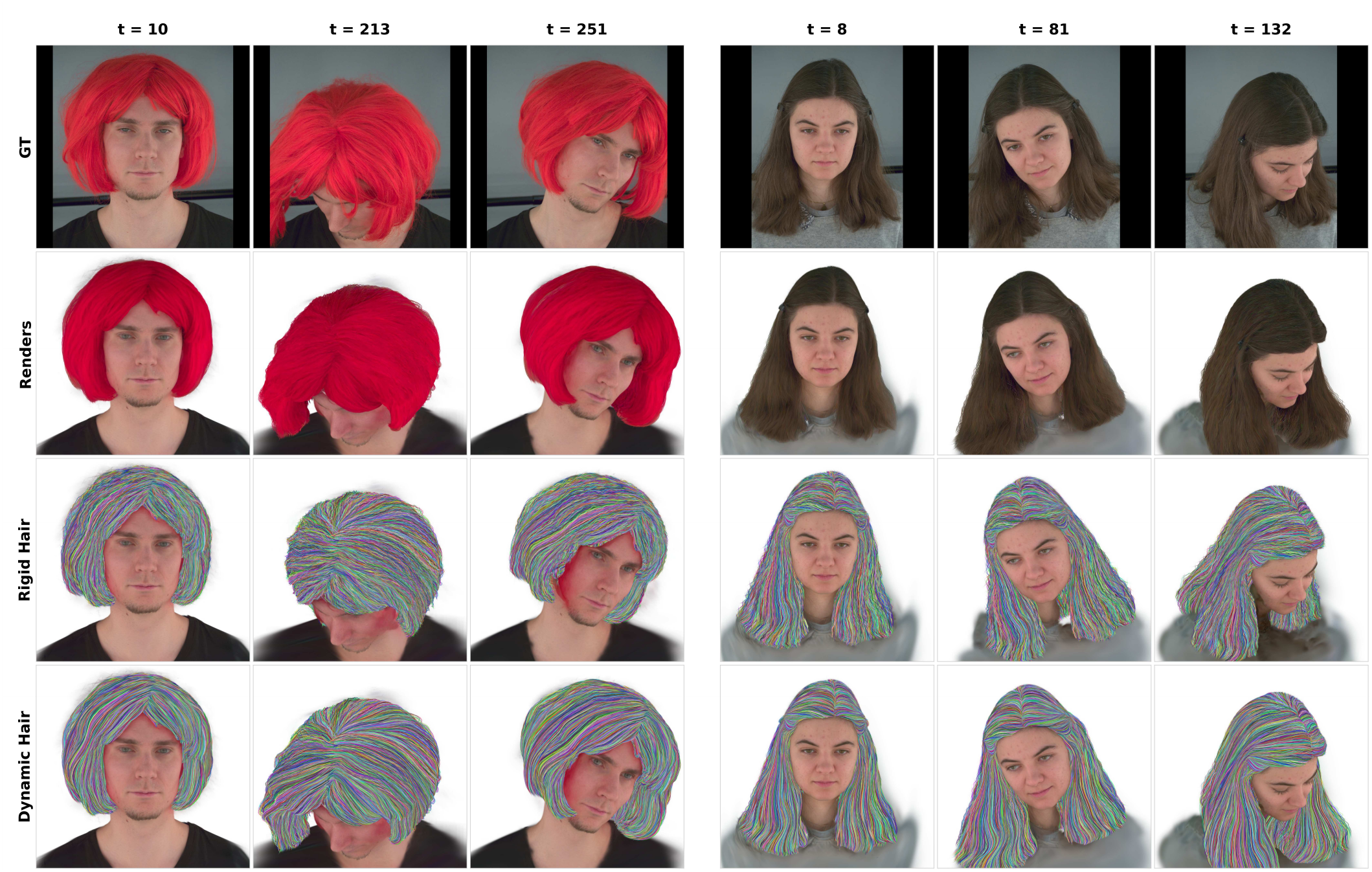}
    \caption{
We show more results on newly captured sequences with diverse hair motions. Notably, ours recovers per-strand temporal hair motion. 
We include an ablation without the non-rigid deformation module, where hair moves rigidly with the head transform.
    }
    \label{fig:hairstyle_tracking}
\end{figure*}

\subsection{Ablation Study}
\cref{tab:ablation} presents an ablation of our key design choices. 
Hair dynamics are evaluated using (1) image-space temporal metrics and (2) motion-based metrics, including velocity explosion ratio, curvature temporal smoothness, displacement spike, and drift.

\smallskip
\noindent\textbf{Visual metrics.} For visual 2D metrics, we compute IoU$_{\text{hair}}$, which measures the overlap between predicted and ground-truth hair silhouettes, and tIoU$_{\text{hair}}$, which measures temporal stability by computing the overlap between predicted hair masks in consecutive frames. 
The ground-truth tIoU$_{\text{hair}}$
averaged across all scenes is 0.951. We also include LPIPS in hair region and $\mathrm{tLPIPS}_{\text{ex}}$.

\smallskip
\noindent\textbf{Physical realism.} To evaluate the realism of hair dynamics beyond standard image-quality metrics, we introduce four motion-based metrics. 
Let $p_{ij}^t \in \mathbb{R}^3$ be the position of the $j$-th point on the $i$-th strand at frame $t$, 
and $\Delta t$ the time step. Then \textit{velocity explosion ratio (VER)} is defined as the ratio of maximum to median speed across all points and frames:
\begin{equation}
    \text{VER} = \frac{\max_{t,i,j} \|v_{ij}^t\|}{\operatorname{median}_{t,i,j} \|v_{ij}^t\| + \epsilon},   
    \quad v_{ij}^t = \frac{p_{ij}^{t+1} - p_{ij}^t}{\Delta t}.
\end{equation}

We also include \textit{Curvature Temporal Smoothness (CTS)}, which quantifies temporal jitter in strand curvature by averaging the frame-to-frame change in curvature vectors across all frames, strands, and strand points. Lower values indicate smoother, more temporally consistent hair deformation. 

To detect sudden, physically implausible positional jumps, we introduce the \textit{Displacement Spike (P95-Disp)} metric, which is defined as 
the 95th percentile of per-frame maximum strand displacements:
\begin{equation}
    \text{P95-Disp} = \operatorname{P}_{95}\!\left(\left\{\max_{i,j} \|p_{ij}^{t+1} - p_{ij}^t\|\right\}_{t=1}^{T-1}\right).
\end{equation}

Finally, \textit{Angular Momentum Drift (AMD)} measures violations of angular momentum conservation. In an isolated system, angular momentum should remain constant over time:
\begin{equation}
    \text{AMD} = \frac{1}{T-1}\sum_{t=1}^{T-1}\|L^t - L^1\|, 
    \quad L^t = \sum_{i,j} \left(p_{ij}^t - \bar{p}^t\right) \times v_{ij}^t,
\end{equation}
where $\bar{p}^t = \frac{1}{NM}\sum_{i,j}p_{ij}^t$ is the center of mass. 
For all metrics, lower values indicate more physically plausible motion.

In~\cref{tab:ablation}, we compute all metrics for the same number of iterations (240k), averaged across 2 scenes with 3 motion types: nodding, left-right motion, 
and head rotation. Maya-simulated strands are included for reference. While Maya simulations expectedly produce better physics metrics (except VER), we observe degradation in coverage metrics due to suboptimal parameters.
We study the effect of each conditioning signal in the hair deformation model. Conditioning on absolute pose worsens generalization, while removing gravity prevents strands from falling properly (e.g., during nodding). Removing acceleration (``w/o acc'') slightly degrades the metrics and reduces motion expressiveness.
We find that the deformation model plays a crucial role. Without it, hair motion appears rigid and unrealistic (see ``Rigid Hair'' vs. ``Dynamic Hair'' in~\cref{fig:hairstyle_tracking}). Replacing the LSTM head pose encoder with an MLP results in stiffer motion (see negative $\mathrm{tLPIPS}_{\text{ex}}$).
Modulating strand features with head pose features improves performance across all metrics (compare ours with ``w/o film'').
Removing \textit{$\mathcal{L}_\text{elastic}$} significantly degrades  results. 
Strands tend to stretch, reducing physical plausibility.
See more ablations in \suppl.

\begin{figure}[t]
    \centering
    \setlength{\tabcolsep}{1pt}
    \renewcommand{\arraystretch}{0.5}

\makebox[0.2\textwidth]{\scriptsize Reconstructed hairstyle}%
\makebox[0.2\textwidth]{\scriptsize Brightness$\uparrow$}%
\makebox[0.2\textwidth]{\scriptsize Brightness $\downarrow$}%
\makebox[0.2\textwidth]{\scriptsize Brightness $\downarrow$ + trim (20 \%)}
\includegraphics[width=0.8\textwidth]{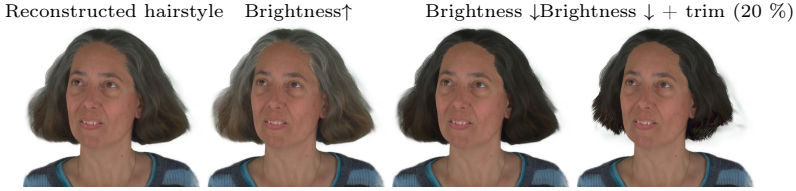}
    \caption{
        \textbf{Hairstyle manipulation.} The reconstructed dynamic hairstyle can be further manipulated by changing the brightness of Gaussians or trimming them.
    }
    \label{fig:editing}
\end{figure}

\begin{table}[t]
\centering
\caption{Quantitative comparison on physical realism of hair motion. Results are averaged across 2 scenes and 3 motion types, estimated at the same number of iterations.}
\label{tab:ablation}
\setlength{\tabcolsep}{4pt}
\resizebox{0.9\columnwidth}{!}{%
\begin{tabular}{l|cccc|cccc}
\multirow{2}{*}{Method}

& \multicolumn{4}{c|}{Silhouette \& Appearance}
& \multicolumn{4}{c}{Physics} \\

& IoU$_{\text{hair}}\uparrow$
& tIoU$_{\text{hair}}\uparrow$
& LPIPS$_{\text{hair}}\downarrow$
& tLPIPS$_{\text{ex}}$
& VER$\downarrow$
& P95$\downarrow$
& CTS$\downarrow$
& AMD$\downarrow$ \\
\hline

Ours             & 0.883 & 0.945 & 0.101 & 0.0024 & 26.88 & 0.071 & 0.00022 & 7787 \\
Maya (Reference)          & 0.789	&0.941& 	0.131 	&0.0080 & 33.73 & 0.057 & 0.00010 & 7120 \\
\hline
w/ pose        & 0.885 & 0.945 & 0.101 & 0.0028 & 36.27 & 0.071 & 0.00019 & 7816 \\
w/o acc.   & 0.878 & 0.944 & 0.101 & 0.0026 & 29.05 & 0.077 & 0.00021 & 8332 \\
w/o $\mathbf{g}$        & 0.852 & 0.932 & 0.104 & 0.0058 & 23.13 & 0.070 & 0.00022 & 8159 \\
\hline
w/o $\mathcal{L}_\text{elastic}$      & 0.873 & 0.943 & 0.103 & 0.0047 & 283.75 & 1.159 & 0.00156 & 196993 \\
w/o film         & 0.877 & 0.945 & 0.102 & 0.0009 & 40.46 & 0.104 & 0.00038 & 8277 \\
w/ mlp enc.            & 0.879 & 0.949 & 0.102 & -0.0010 & 32.18 & 0.065 & 0.00020 & 7377 \\

\hline
\hline
\end{tabular}}
\end{table}

\section{Discussion}
\noindent\textbf{Applications.} Disentangled modeling of head and hairstyle provides rich possibilities for direct hair manipulation. For example, we can edit the brightness of Gaussians or trim the hairstyle length while preserving realistic hair motion dynamics (see~\cref{fig:editing}). Furthermore, the explicit strand-based representation allows us to derive physically meaningful motion signals from the reconstructed hair geometry, see~\cref{fig:hairstyle_tracking}, which can be used for transfer applications.

~

\noindent\textbf{Limitations.}
Our method models only hair–head interactions; hair–body collisions are handled implicitly through data-driven priors and are not explicitly resolved. The approach depends on hair silhouette estimates from pre-trained segmentation models, which can be inaccurate for complex hairstyles or under occlusion. We also observe that rendering quality at hair–head boundaries can exhibit artifacts due to the discrete transition between the two Gaussian representations. Additionally, reliance on the landmark detector and tracker can lead to incorrect reconstructions when hair occludes the face. Our formulation provides a foundation for future large-scale learning of dynamic hair, where richer datasets and explicit collision modeling can be naturally integrated. 

~

\noindent\textbf{Conclusion.}
In this work, we presented a novel dynamic head avatar framework with explicit strand-based hair, learned end-to-end from multi-view video. Our method represents hair as strand-aligned 3D Gaussians whose deformations are predicted by a temporal motion model conditioned on head angular velocity, acceleration, and relative gravity. By combining an LSTM head pose encoder with FiLM modulation and physics-based regularization, the approach produces temporally coherent hair dynamics with strand-level geometric fidelity. Experiments demonstrate high visual quality and improved motion plausibility compared with state-of-the-art head avatar methods, both quantitatively and qualitatively. We believe that explicitly modeling hair structure and dynamics within a unified Gaussian avatar framework is a step toward more realistic and controllable digital humans.

\section{Acknowledgements}
Vanessa Sklyarova is supported by the Max Planck ETH Center for Learning Systems.
Justus Thies is supported by the ERC Starting Grant 101162081 ``LeMo'' and the DFG Excellence Strategy - EXC-3057.
This work is supported by the ERC Consolidator Grant Gen3D (101171131).  
Berna Kabadayi is supported by the IMPRS-IS.
Gerard Pons-Moll is a member of the ML Cluster of Excellence, EXC-2064/1 – Project No. 390727645, and is endowed by the Carl Zeiss Foundation.
This work is supported by the BMBF: Tübingen AI Center, FKZ: 01IS18039A, and by the DFG – 409792180.
This work is supported by the Swiss National Science Foundation Advanced Grant 216260: ``Beyond Frozen Worlds: Capturing Functional 3D Digital Twins from the Real World.''

The authors would like to thank Zhanfeng Liao for providing data from HHAvatar, Egor Zakharov for supervising Haonan during his time at ETH, and Benjamin Pellkofer for IT support.

\smallskip

\noindent\textbf{Disclosure.} While MJB is employed by Epic Games, this work was performed solely at, and funded solely by, the Max Planck Society.

\bibliographystyle{splncs04}
\bibliography{abbr, main}

\clearpage

\appendix

\section{Background}
\subsection{Dynamic Head Avatar} 

We model the expression- and pose-dependent head and upper body following the GHA method~\cite{xu2023gaussianheadavatar}.
Specifically, we adopt a geometry-guided initialization strategy based on an implicit SDF and Deep Marching Tetrahedra to initialize the canonical representation of 3D Gaussians with dynamic network weights, which improves the stability and convergence of training. During each iteration, three pairs of specialized MLPs predict position displacement, appearance, and Gaussian attributes.
For each pair of MLPs, the individual predictions conditioned on pose parameters and expression parameters are subsequently fused using spatially-varying blending weights $w^\text{exp}_k$, $w^\text{pose}_k$, which are computed based on the proximity of each point to 3D facial landmarks. 
Those predictions are used to further refine the canonical state for the non-rigid deformed Gaussian representation. Finally, a global rigid transformation is applied to project the deformed Gaussians into the world coordinate system.

~

\noindent\textbf{Canonical representation of the head.}
Following~\cite{xu2023gaussianheadavatar}, we first train a mesh-based head model that learns neural geometry and appearance fields on a tetrahedral grid conditioned on BFM~\cite{paysan20093d} parameters. 
From the converged mesh head model, we extract a canonical set of $K$ 3D Gaussian primitives.
Each primitive is parameterized by its position $\mathbf{x}_k \in \mathbb{R}^3$, a learned feature vector $\mathbf{f}_k \in \mathbb{R}^{128}$, log-scale $\mathbf{s}_k \in \mathbb{R}^3$, rotation quaternion $\mathbf{q}_k \in \mathbb{R}^4$, and logit-opacity $o_k \in \mathbb{R}$. 
To improve mouth interior quality, we augment the canonical set with additional Gaussians that are randomly sampled around mouth landmarks.

~

\noindent\textbf{Deformation module.}
Two separate MLPs predict per-point displacements driven by facial expression and head pose for each Gaussian primitive.
Specifically, the expression deformation MLP $f^\text{exp}_\text{def}$ takes as input the positional encoding $\gamma(\cdot)$ of canonical coordinates concatenated with the expression coefficient vector $\mathbf{e} \in \mathbb{R}^{64}$:
\begin{equation}
    \Delta \mathbf{x}^\text{exp}_k
    =
    f^\text{exp}_\text{def}
    \!\left(
    \gamma(\mathbf{x}_k), \mathbf{e}
    \right).
\end{equation}
Similarly, the pose deformation MLP $f^\text{pose}_\text{def}$ takes as input canonical coordinates and the pose $\mathbf{p} \in \mathbb{R}^{6}$ (global translation and rotation): 
\begin{equation}
    \Delta \mathbf{x}^\text{pose}_k
    =
    f^\text{pose}_\text{def}
    \!\left(
    \gamma(\mathbf{x}_k), \gamma(\mathbf{p})
    \right).
\end{equation}
The two deformations are fused using spatially varying weights determined by proximity to 3D facial landmarks. 
Let $d_k$ be the distance from the point $k$ to its nearest landmark. We define blending weights as: 
\begin{equation}
    w^\text{exp}_k = \mathrm{clamp}\!\left(\frac{d_\text{far} - d_k}{d_\text{far} - d_\text{near}},\, 0,\, 1\right), \quad w^\text{pose}_k = 1 - w^\text{exp}_k,
\end{equation}
where $d_\text{near}$ and $d_\text{far}$ define a transition region. Points near landmarks (small $d_k$) are primarily expression-driven, while distant points follow the global head pose.
The final deformed position is given by:
\begin{equation}
    \mathbf{x}'_k = \mathbf{x}_k + \alpha\!\left(w^\text{exp}_k \cdot \Delta \mathbf{x}^\text{exp}_k + w^\text{pose}_k \cdot \Delta \mathbf{x}^\text{pose}_k\right),
\end{equation}
where $\alpha$ is a predefined deformation scale factor that prevents large displacements.

~

\noindent\textbf{Appearance and attributes.}
Two additional pairs of MLPs take per-point learned features $\mathbf{f}_k$ as inputs and predict per-point appearance and Gaussian attribute offsets.
The \emph{color MLPs} take $(\mathbf{f}_k, \mathbf{e})$ and $(\mathbf{f}_k, \gamma(\mathbf{p}))$ as inputs for the expression- and pose-conditioned branches, respectively, producing 32-channel color features $\mathbf{c}^{exp}_k$ and $\mathbf{c}^{pose}_k$ that are blended with the same landmark-based weights $w^\text{exp}_k$, $w^\text{pose}_k$.
The final color $\mathbf{c}'_k$ is obtained by:
\begin{equation}
    \mathbf{c}'_k = w^\text{exp}_k \cdot \mathbf{c}^{exp}_k + w^\text{pose}_k \cdot \mathbf{c}^{pose}_k .
\end{equation}
The \emph{attribute MLPs} share the same inputs and predict 8-dimensional offsets $\delta_k = [\delta\mathbf{s}_k, \delta\mathbf{q}_k, \delta o_k]$ (3 for scale, 4 for rotation, 1 for opacity), also blended via landmark weights.
The final Gaussian attributes are obtained by applying the scaled offsets to the canonical values:
\begin{equation}
    \mathbf{s}'_k = \exp\!\left(\mathbf{s}_k + \beta\, \delta\mathbf{s}_k\right), \quad
    \mathbf{q}'_k = \mathrm{normalize}\!\left(\mathbf{q}_k + \beta\, \delta\mathbf{q}_k\right), \quad
    o'_k = \sigma\!\left(o_k + \beta\, \delta o_k\right),
\end{equation}
where $\beta$ is an attribute scale factor and $\sigma$ denotes the sigmoid function. 
Lastly, all head Gaussians are transformed to world coordinates using the global rigid pose.

\subsection{Gaussian Splatting for hair rasterization}

To use 3D Gaussian Splatting for soft-rasterization of hair, we force Gaussians to lie on line segments. The mean of each Gaussian is attached to the middle of each line segment $\mu_{ij} = \frac{1}{2} \big( p_{ij} + p_{i,j+1} \big)$.  
The generated Gaussians have a scaling vector $s_{ij}$ that is set to be proportional to the length of the strand, while the other scales are set to small values $\epsilon$: $s_{ij}=\{\frac{1}{2}\|p_{ij+1} - p_{i,j}\|_{2}, \epsilon, \epsilon \}$, where the rotation quaternion for the x-axis is aligned with the segment direction $d_{ij}$.
All opacity values for the hair Gaussians are set to 1.

\section{Training details}

In this section, we provide additional details on data preprocessing, optimization,  datasets, and compute used for training and inference.

\subsection{Optimization details}

\noindent\textbf{Canonical initialization.} For the body, we use the geometry-guided initialization strategy from Gaussian Head Avatar~\cite{xu2023gaussianheadavatar} with default weights and apply it to all timesteps.

To obtain the canonical hairstyle, we incorporate the single-view Im2Haircut~\cite{sklyarova2025im2haircut} prior and optimize it across all available views. To improve the scalp region for strand growth, we discard strand roots outside the hair silhouette. The static reconstruction loss consists of photometric, geometric, and regularization terms:
\begin{equation}
\begin{split}
    \mathcal{L}_\text{static} = {} & \lambda_\text{rgb}\mathcal{L}_\text{rgb} + \lambda_\text{ssim}\mathcal{L}_\text{ssim} + \lambda_\text{vgg}\mathcal{L}_\text{vgg} \\
    & + \lambda_\text{seg}\mathcal{L}_\text{seg} + \lambda_\text{orient}\mathcal{L}_\text{orient} + \lambda_\text{pen}\mathcal{L}_\text{pen} \\
    & + \lambda_\text{shape}\mathcal{L}_\text{shape} + \lambda_\text{smooth}\mathcal{L}_\text{smooth} + \lambda_\text{length}\mathcal{L}_\text{length} + \lambda_\text{pca}\mathcal{L}_\text{pca} ,
\end{split}
\end{equation}
where $\mathcal{L}_\text{rgb}$ is the L1 photometric loss, $\mathcal{L}_\text{ssim} = 1 - \text{SSIM}(\hat{I}, I)$ is the structural similarity loss, and $\mathcal{L}_\text{vgg}$ is the LPIPS perceptual loss.
$\mathcal{L}_\text{seg}$ supervises the rendered hair silhouette using an asymmetric recall loss with exponential decay during training.
$\mathcal{L}_\text{orient}$ matches predicted hair strand orientations to ground truth orientation maps using a circular distance formulation.
$\mathcal{L}_\text{pen}$ prevents strand--head penetration by penalizing the distance from strand points detected inside the FLAME mesh to its surface.
To further increase smoothness due to sparse views with limited coverage, we introduce spatial regularization losses. Let $\mathcal{S}_i = \{p_{i1}, \dots, p_{iL}\}$ denote strand $i$ with $L$ points, and let $\mathcal{N}_k(i)$ denote its $k$ nearest neighbors based on root position ($k{=}4$).

~

\noindent
\textit{Shape consistency.} $\mathcal{L}_\text{shape}$ penalizes curvature signature differences between neighboring strands. We first compute a per-strand curvature descriptor as the sequence of bending angles between consecutive edges:
\begin{equation}
    \kappa_{im} = \arccos\left(\frac{(p_{i,m+1} - p_{i,m}) \cdot (p_{i,m+2} - p_{i,m+1})}{\|p_{i,m+1} - p_{i,m}\| \, \|p_{i,m+2} - p_{i,m+1}\|}\right),
\end{equation}
yielding a descriptor $\boldsymbol{\kappa}_i = (\kappa_{i1}, \dots, \kappa_{i,L-2})$ per strand. The loss is:
\begin{equation}
    \mathcal{L}_\text{shape} = \frac{1}{Nk} \sum_{i=1}^{N} \sum_{j \in \mathcal{N}_k(i)} \|\boldsymbol{\kappa}_i - \boldsymbol{\kappa}_j\|^2 .
\end{equation}

~

\noindent
\textit{Strand smoothness.} $\mathcal{L}_\text{smooth}$ encourages neighboring strands to have similar shapes by comparing root-normalized geometry:
\begin{equation}
    \mathcal{L}_\text{smooth} = \frac{1}{Nk} \sum_{i=1}^{N} \sum_{j \in \mathcal{N}_k(i)} \sum_{l=1}^{L} \|(p_{il} - p_{i1}) - (p_{jl} - p_{j1})\|^2 .
\end{equation}

~

\noindent
\textit{Length smoothness.} $\mathcal{L}_\text{length}$ enforces similar total arc lengths among neighboring strands. Denoting the total length of strand $i$ as $\ell_i = \sum_{m=1}^{L-1} \|p_{i,m+1} - p_{i,m}\|$:
\begin{equation}
    \mathcal{L}_\text{length} = \frac{1}{Nk} \sum_{i=1}^{N} \sum_{j \in \mathcal{N}_k(i)} (\ell_i - \ell_j)^2 .
\end{equation}

~

\noindent
\textit{PCA consistency.} $\mathcal{L}_\text{pca}$ regularizes strands on the back of the head (which receive less multi-view coverage) by matching their PCA coefficient distribution to that of front-view strands. Let $\boldsymbol{\mu}^f, \boldsymbol{\sigma}^f$ and $\boldsymbol{\mu}^b, \boldsymbol{\sigma}^b$ denote the per-component mean and variance of PCA coefficients for front and back strands, respectively. The loss applies an exponentially decaying weight $w_d = \exp(-0.05\,d)$ over PCA dimension $d$:
\begin{equation}
    \mathcal{L}_\text{pca} = \frac{1}{D}\sum_{d=1}^{D} w_d \left(|\mu^f_d - \mu^b_d| + \tfrac{1}{2}|\sigma^f_d - \sigma^b_d|\right) .
\end{equation}
The loss weights are set as follows: $\lambda_\text{rgb}$=0.2, $\lambda_\text{ssim}{=}0.02$, $\lambda_\text{vgg}{=}0.01$, $\lambda_\text{seg}{=}1.0$, $\lambda_\text{orient}{=}0.5$, $\lambda_\text{pen}{=}20.0$, $\lambda_\text{shape}{=}\lambda_\text{smooth}{=}\lambda_\text{length}{=}0.001$, and $\lambda_\text{pca}{=}0.0001$. The prior learning rates are set to $10^{-5}$ for both the coarse and fine models.
We optimize for 1200 iterations with 4 views for HHAvatar~\cite{Liao2025hhavatar} scenes and for 5000 iterations for our captured scenes, which have static 360$^\circ$ coverage. This yields the canonical hairstyle $\mathcal{H}_0$ used as input for the dynamic stage.
The obtained hair strands have 200 points.

~

\noindent\textbf{Dynamic training stage.} At the beginning of the dynamic stage, we resample the hairstyle to ensure that each strand contains 40 points. Since predicting shifts for each strand point introduces high degrees of freedom that can lead to spikes and zigzag artifacts, we apply strand smoothing for interior points using a 1D averaging filter on each   strand at every iteration. Specifically, each interior point is replaced by the average of itself and its two neighbors along the strand. This acts as a spatial low-pass filter along the strand curve, ensuring   that the deformed hair strands remain smooth polylines rather than exhibiting jagged kinks. The operation is lightweight (a single pass with a 3-point average) and fully differentiable, so gradients flow through it during training.

We optimize the dynamic model for 320,000 iterations. The loss weights are set as follows: $\lambda_\text{rgb}{=}2.0$, $\lambda_\text{ssim}{=}0.2$, $\lambda_\text{vgg}{=}0.1$, $\lambda_\text{seg}{=}10.0$ (with exponential decay), $\lambda_\text{orient}{=}1.0$, $\lambda_\text{penetr}{=}1.0$, $\lambda_\text{elastic}{=}500.0$, and $\lambda_\text{cg}{=}\lambda_\text{cs}{=}\lambda_\text{cc}{=}0.01$ for the three color regularization terms.
We increase the VGG weight to $\lambda_\text{vgg}{=}0.6$ for the last 80,000 iterations to enhance the appearance.

For hair rasterization, we represent appearance using spherical harmonics of degree 3 (SH=3) and fix the opacity to 1.

\subsection{Data preprocessing}

\textit{Head tracking:}
To obtain consistent head tracking from multi-view images, we use a BFM-based~\cite{paysan20093d} multi-view 3DMM fitting method\footnote{\url{https://github.com/YuelangX/Multiview-3DMM-Fitting}}. 
The tracker estimates identity, expression, and pose parameters of the face by fitting the Basel Face Model to all views simultaneously, producing temporally and geometrically consistent face alignment across cameras.

~

\noindent\textit{Segmentation masks:}
For image preprocessing, we extract segmentation masks for the body using MODNet~\cite{MODNet}, which provides robust human matting and foreground segmentation. 
Hair masks are predicted using CDGNet~\cite{CDGNET}, a semantic segmentation network specialized for hair region detection.

~

\noindent\textit{Hair orientation maps:}
To estimate local hair directions, we compute orientation maps using a bank of Gabor filters applied to the input images. 
Following prior work~\cite{Sklyarova2023NeuralHP, Wu2024monohair, Luo2024gaussianhair, zakharov2024gh} on hair reconstruction, the filter bank spans multiple orientations, and the dominant response at each pixel is used to estimate the local hair strand direction within the hair mask. 
We additionally use the preprocessing scripts provided by Im2Haircut~\cite{sklyarova2025im2haircut} to generate the input representation required by the static hair prior model.

\subsection{Datasets}

\noindent\textit{HHAvatar dataset:}
We show results on 3 scenes from the HHAvatar~\cite{Liao2025hhavatar} dataset, which includes videos with 4 synchronized cameras, with around 3 minutes per scene. We use a 60\%/40\% train–test split for the black and yellow hair scenes from HHAvatar~\cite{Liao2025hhavatar}, and an 80\%/20\% split for the brown hair scene.

~

\noindent\textit{New dataset:}
For our new dataset, we capture subjects using 15 synchronized multi-view video cameras with an angular coverage of 93$^\circ$ left-to-right and 32$^\circ$ up-to-down. The sequences are recorded at a resolution of 7.1 MP and 72 fps, resulting in approximately three minutes of footage per subject. In addition, we capture a full 360$^\circ$ spinning sequence of each subject both with and without a hair cap, enabling reconstruction of otherwise occluded regions such as the ears and neck, and improving hairstyle reconstruction quality. The dataset includes 10 recordings focused on facial expressions and 10 recordings containing diverse hair motions. These sequences cover a wide range of motions, including emotional expressions (e.g., shout + laugh, angry + sad), isolated facial articulations (eyes, mouth, jaw, and tongue), and head movements such as rotation, nodding, tilting, impulse motion, and horizontal figure-eight trajectories. Additionally, the dataset contains dedicated test sequences for hair-only motion, expression-only motion, and joint motion involving both hair dynamics and facial expressions.

\subsection{Compute time}

Static hairstyle optimization takes 24~min on HHAvatar sequences and 100~min on newly-captured data. MeshHead initialization takes $\sim$3--4 hours, and full-model training takes $\sim$5 days. Model inference runs at ${\sim}21$~FPS (23~FPS with all data preloaded on GPU) at $1024 \times 1024$ resolution on an A100.

\section{Evaluation details}

In this section, we provide more details on the baselines and metrics used for evaluation of our method.

\subsection{Baselines}

\textit{DynHair} is a person-specific method: for each subject, the model is trained on multi-view video capturing diverse head motions, and at test time generalizes to novel motions of that same subject. Concurrent works on dynamic hair modeling, such as HADES \cite{Liao_2025_HADES} and HHAvatar \cite{Liao2025hhavatar}, do not provide publicly available implementations, making direct comparison difficult. Therefore, we instead compare with dynamic head avatar methods, including Gaussian Head Avatar (GHA) \cite{xu2023gaussianheadavatar} and GaussianAvatars (GA) \cite{qian2024gaussianavatars}, as well as a physics-based simulator. All methods are evaluated on the same subjects, input videos, and train/test splits. 

We launch Gaussian Head Avatar~\cite{xu2023gaussianheadavatar} and GaussianAvatars~\cite{qian2024gaussianavatars} using the code from their official repositories. 
We train each baseline for around 600,000 iterations. 
For comparison with the physics-based simulator, we use Maya~\cite{maya} due to its flexibility in exporting 3D positions.
We simulate 200 sparse strands (16 points each) and interpolate to 40 points for each strand and to the original hairstyle using the k-nearest neighbors with $k=10$.
The parameters used for dynamic hair simulation are as follows:
selfCollide $=1$, friction $=0.51087$, stickiness $=0.51087$,
stretchResistance $=600$, compressionResistance $=600.0$,
bendResistance $=60$, twistResistance $=1.718$,
extraBendLinks $=3$, mass $=2$, drag $=0.65$,
tangentialDrag $=0.0965909$, damp $=0.8$, stretchDamp $=1$,
dynamicsWeight $=1$, staticCling $=0.025$, and startCurveAttract $=0.6$.

\subsection{Metrics}

For quantitative evaluation in Table 1 of the main paper, we use 229 frames × 4 views for the brown scene, 400 frames × 4 views for the yellow scene, and 280 frames × 4 views for the black scene, all taken from the test split. For FID calculation, we compute the distance between distributions on joint datasets.  

In Table 1 of the main paper, we compute the velocity and acceleration of Gaussians that belong to hair regions across all methods. As point correspondences are not guaranteed for methods that use unstructured Gaussians, we compute per-frame velocity based on mean nearest-neighbour displacements. The acceleration is computed as the finite difference between velocities, which is defined as $a_{t}=\left|v_{t+1}-v_{t}\right |, \quad t = 0, \dots, T-3$. In this case, lower acceleration gives better results, because it measures temporal smoothness, not physical realism. $a_{t}$ shows how abruptly the mean hair displacement changes with high values indicating sudden jumps, flickering, and inconsistencies in temporal hair motion. For fair comparison, we use the same number of Gaussians for metric calculation defined by minimum number of Gaussians across all methods.

In Table 2 of the main paper, we calculate \textit{Curvature Temporal Smoothness (CTS)} metric that quantifies temporal jitter in strand curvature and is defined as:
\begin{equation}
    \text{CTS} = \frac{1}{(T{-}1)N(M{-}2)} \sum_{t,i,m} \|\boldsymbol{\kappa}_{im}^{t+1} - \boldsymbol{\kappa}_{im}^t\|.
\end{equation}

\section{Applications}

\noindent\textbf{Temporal Strand Correspondence.} Our hair representation defines strands in a shared canonical space, where each strand is deformed per-frame by a deformation network that is conditioned on head velocity, acceleration, and gravity. As a result, strand correspondence across frames is established by construction: strand $i$ at frame $t$ corresponds to the same canonical strand at any other frame $t'$. This consistent indexing enables  temporally coherent hair editing: any per-strand operation, such as trimming, thinning, or recoloring, defined at a single frame automatically propagates to the entire sequence without requiring explicit per-frame user annotation. 

~

\noindent\textbf{Editing.}
We apply hairstyle edits as a post-processing step on the deformed hair Gaussians. At each frame, the full hair strand representation is first passed through the learned deformation network to obtain pose- and   expression-dependent Gaussian attributes (position, color, scale, rotation, opacity). Edits are then applied to the resulting Gaussians before rendering. Each strand $i$ consists of $L$ Gaussians $\{p_{ij}\}_{j=1}^{L}$ ordered from root to tip.   

~

\noindent\textbf{Hairstyle color.} Color adjustment is performed by uniformly scaling the RGB color attributes of all hair Gaussians by a user-specified factor $\alpha \in (0, 1)$, i.e., $\mathbf{c}_{ij}\leftarrow \alpha \cdot \mathbf{c}_{ij}$. For example, $\alpha = 0.7$ reduces brightness by 30\%. This preserves the relative color variation and shading across the hair while uniformly darkening the overall appearance.        

~

\noindent\textbf{Trimming.} Given a user-specified keep ratio $r \in (0, 1]$, we set the opacity of the tail Gaussians $\{p_{ij}\}_{j=\lfloor r \cdot L \rfloor}^{L}$ to zero, effectively shortening the visible    hair while preserving the natural deformation of the remaining portion. This post-deformation editing strategy ensures that the deformation network operates on the same strand topology it was trained on.

~

\section{Additional experiments}
In this section, we provide additional experiments on baseline comparisons and an extended ablation study.
\subsection{Comparison with baselines}
\subsubsection{User study.}
To further evaluate our method, we conducted a user study with 16 participants on 9 test sequences of hair motion, shown in ~\cref{tab:user_study}.
Participants were asked to evaluate three aspects: (1) \textbf{Appearance Quality}: ``Which rendering looks more realistic, especially in the hair region?''; (2) \textbf{ Hair Motion: } ``Does the hair move and react naturally to physics (e.g., gravity, inertia) and to the character’s head or body motion?''; and (3) \textbf{Overall Preference: }``Which method do you prefer overall?''
The participants found that our method has significantly better appearance quality. 
It exhibits less jitter and follows the head motion more accurately than the physics simulator~\cite{maya} and GHA~\cite{xu2023gaussianheadavatar}.

\begin{table}[t]
\centering
\caption{\textbf{User study results.} Our method outperforms the baselines in appearance quality, hair motion, and overall preference. }
\setlength{\tabcolsep}{10pt} %
\begin{tabular}{lccc}
\toprule
Method & Appearance Quality & Hair Motion & Overall Preference \\
\midrule
GHA~\cite{xu2023gaussianheadavatar}  & 9.72\%  & 9.72\%  & 11.11\% \\
Maya~\cite{maya} & 22.22\% & 32.64\% & 23.61\% \\
Ours & \textbf{68.06\%} & \textbf{57.64\%} & \textbf{65.28\%} \\
\bottomrule
\end{tabular}
\label{tab:user_study}
\end{table}

\subsubsection{Comparison on new scenes.}

In~\cref{tab:comparison_new_scenes}, we evaluate metrics for comparison with GHA~\cite{xu2023gaussianheadavatar} on 2 newly captured test sequences. We report results averaged across all available cameras as well as on a held-out view. For reference, tIoU$_{\text{gt}}$=0.933 and tIoU$_{\text{gt}}$=0.902 on scenes 603 and 604 self-reenactment, and tIoU$_{\text{gt}}$=0.937 and tIoU$_{\text{gt}}$=0.911 for novel view metrics.

\begin{figure*}[t]
    \centering
    \includegraphics[width=0.99\textwidth]{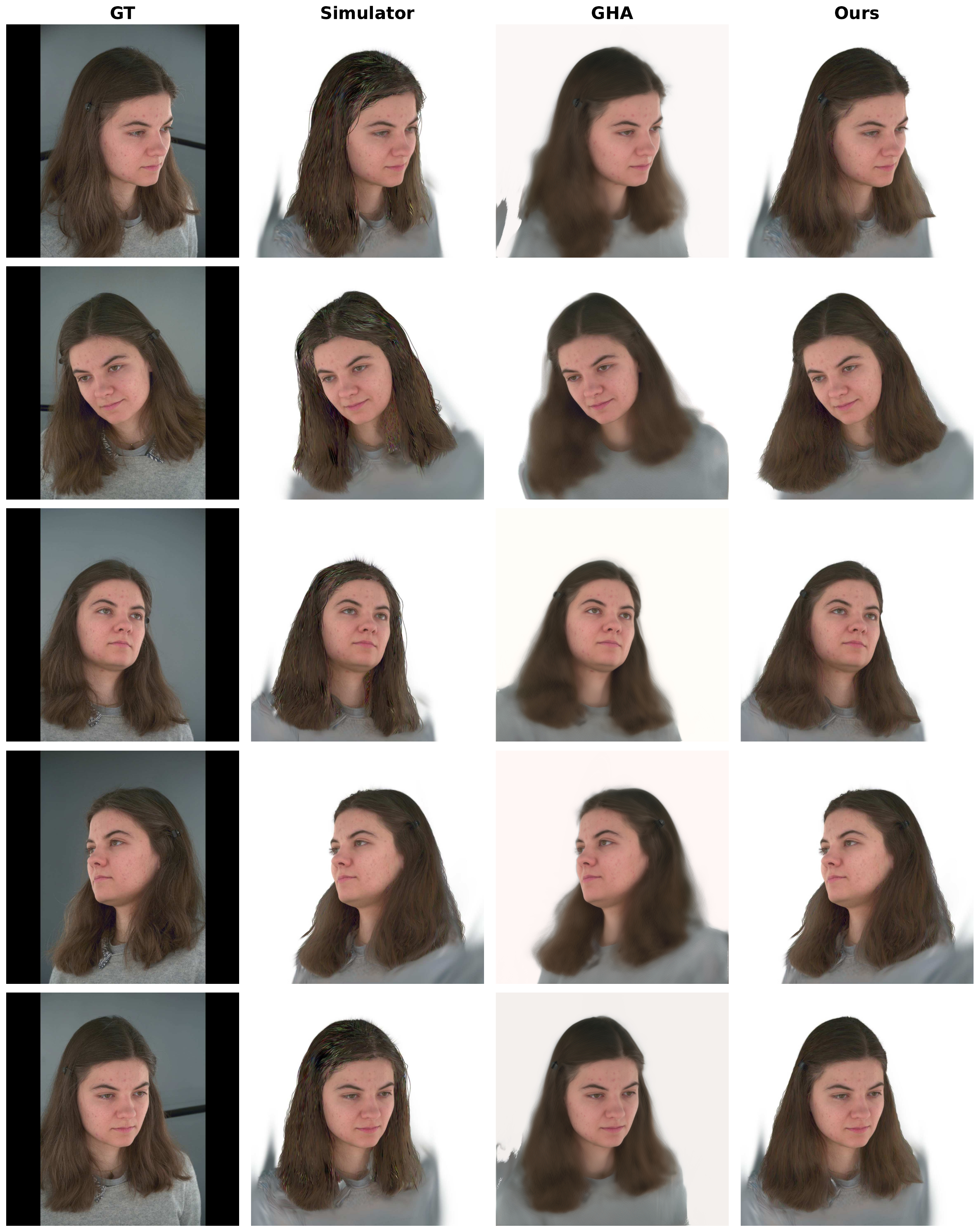}\\[2pt]
    
    \caption{Extended comparison with baselines on new captures.}
    \label{fig:self_reenact_603}
\end{figure*}

\begin{figure*}[t]
    \centering
    \includegraphics[width=0.99\textwidth]{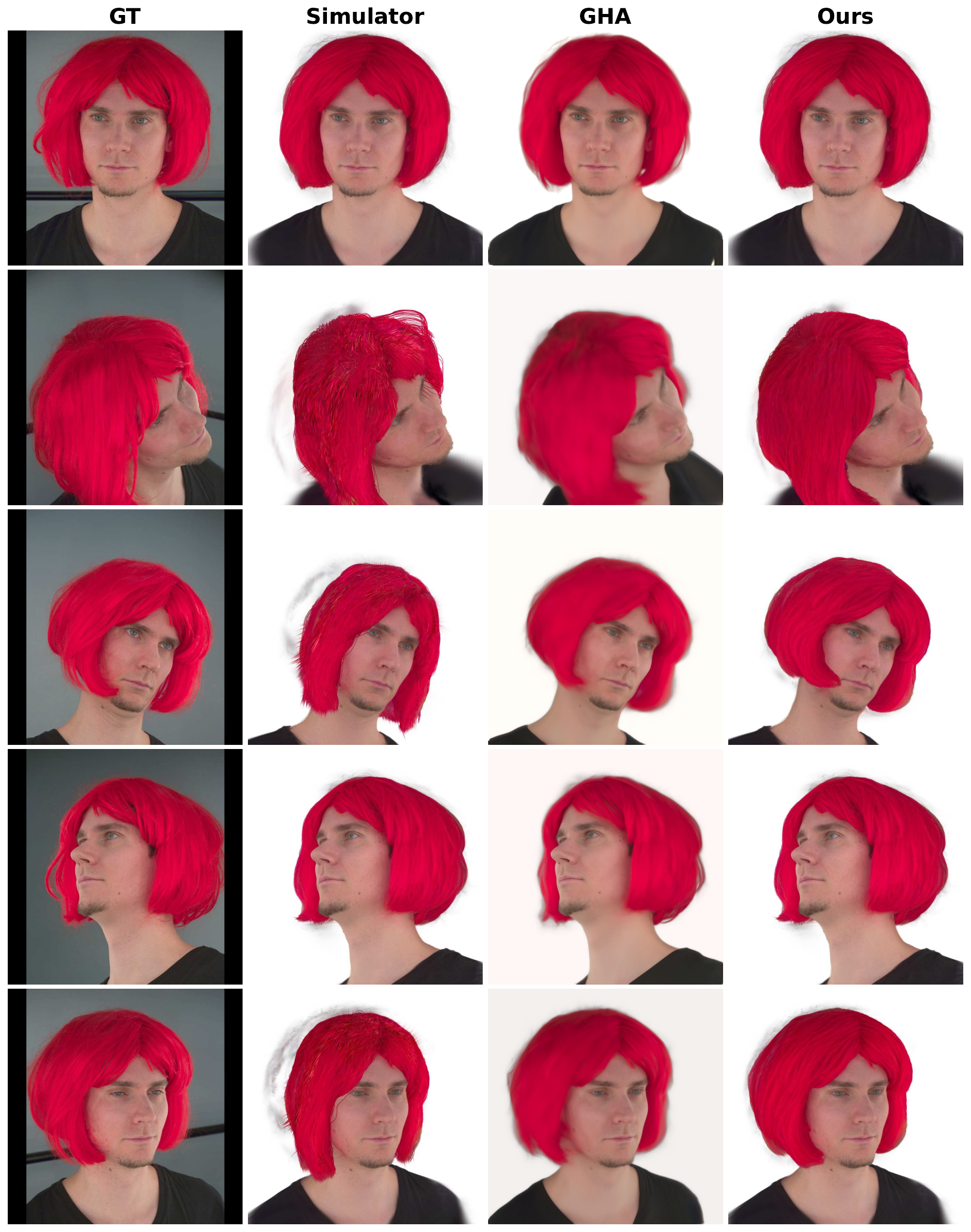}\\[2pt]
    
    \caption{Extended comparison with baselines on new captures.}
    \label{fig:self_reenact_604}
\end{figure*}

In~\cref{fig:self_reenact_603} and~\cref{fig:self_reenact_604}, we present qualitative comparisons with GHA~\cite{xu2023gaussianheadavatar} and simulations from Maya~\cite{maya}. Our method is able to produce finer high-frequency details.  Meanwhile, GHA suffers from blurry results and poor temporal consistency, while Maya has sagging effect artifacts due to manually obtained physical parameters.

\subsection{Extended ablation}

\begin{table*}[t]
\centering
\caption{Ablation study on importance of color regularization, increased weight for the VGG loss and number of views.}
\label{tab:ablation_color}
\setlength{\tabcolsep}{5pt}
\renewcommand{\arraystretch}{1.15}
\resizebox{\textwidth}{!}{%
\begin{tabular}{l|ccc|ccc|c|ccc}
\multirow{2}{*}{Method}
& \multicolumn{3}{c|}{Full Image}
& \multicolumn{3}{c|}{Hair Region}
& Temporal
& \multicolumn{3}{c}{Hair Silhouette} \\
& PSNR$\uparrow$
& SSIM$\uparrow$
& LPIPS$\downarrow$
& PSNR$\uparrow$
& SSIM$\uparrow$
& LPIPS$\downarrow$
& tLPIPS$_{\text{ex}}\downarrow$
& IoU$\uparrow$
& tIoU$_{\text{pred}}\uparrow$
& tIoU$_{\text{gt}}\uparrow$ \\
\hline
Ours & 21.21 & 0.738 & 0.189 & 20.52 & 0.645 & 0.098 & 0.005 & 0.882 & 0.943 & 0.952 \\
w/o increased $\lambda_\text{vgg}$ & 20.88 & 0.753 & 0.195 & 20.01 & 0.670 & 0.100 & 0.003 & 0.881 & 0.943 & 0.952 \\
w/o $\mathcal{L}_\text{color\_reg}$ & 20.95 & 0.737 & 0.189 & 20.27 & 0.644 & 0.098 & 0.006 & 0.884 & 0.943 & 0.952 \\
1 view & 17.35 & 0.706 & 0.250 & 15.80 & 0.600 & 0.113 & 0.003 & 0.763 & 0.939 & 0.946 \\
\hline

\end{tabular}%
}
\end{table*}

\begin{figure*}[t]
    \centering
    \includegraphics[width=0.9\textwidth]{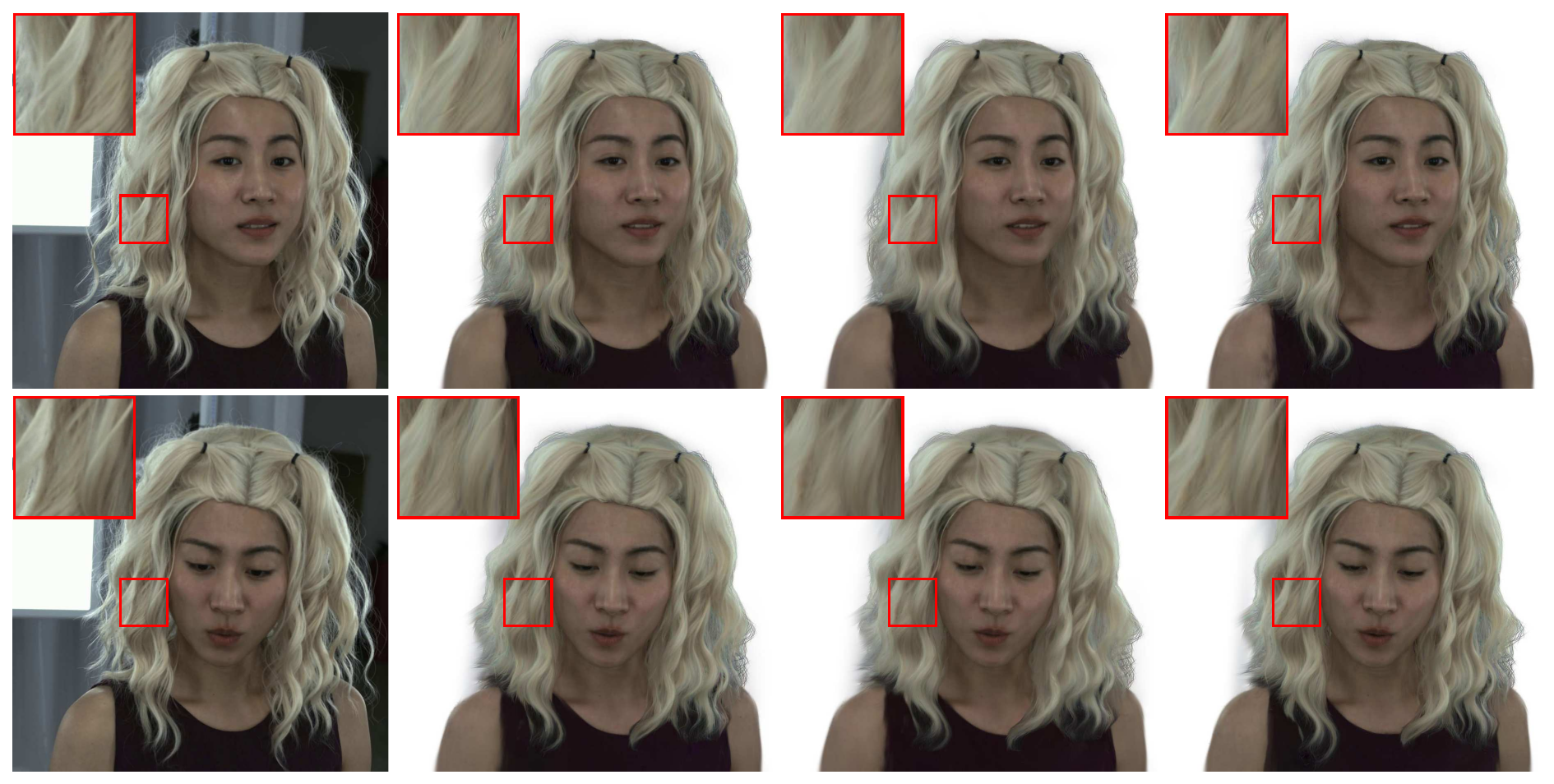}\\[2pt]
\makebox[0.23\textwidth]{\scriptsize GT}%
\makebox[0.23\textwidth]{\scriptsize Ours}
\makebox[0.23\textwidth]{\scriptsize w/o increased $\lambda_\text{vgg}$}%
\makebox[0.23\textwidth]{\scriptsize w/o $\mathcal{L}_\text{color\_reg}$}
\caption{Ablation on color regularization and increased weight for the VGG loss.}
\label{fig:ablation_color}
\end{figure*}

In \cref{tab:ablation_color}, we ablate the importance of color regularization, VGG loss weight, and number of input views. Results are averaged across the test sets of 2 subjects at 320,000 iterations. Omitting the color regularization loss $\mathcal{L}_\text{color\_reg}$ leads to decreased performance across the majority of metrics. Training without an increased $\lambda_\text{vgg}$ also yields worse results. Finally, while our method could operate in a monocular setting, using a single view leads to consistently worse performance. Note that for the monocular experiment, we report metrics only for the view used during training, whereas all other experiments average metrics across all views. We also present a visual ablation in Fig.~\ref{fig:ablation_color}, showing that the results become blurrier without increasing $\lambda_\text{vgg}$. In contrast, removing $\mathcal{L}_\text{color\_reg}$ does not noticeably degrade the quality of visible strands.

\subsection{Limitations}
Despite producing plausible dynamic hair geometry and appearance, our method has several limitations.

\textbf{Hair-face boundary artifacts.} Since hair and face are reconstructed with separate representations, strand-based Gaussians for hair and mesh-anchored Gaussians for the face, visible artifacts can occur at their boundary. The quality of this blending depends on accurate per-frame segmentation and tracking, and errors in either can propagate into the rendered boundary region. Improving the robustness of this blending, for instance through joint segmentation-aware optimization or learned boundary blending, is a promising direction for future work.

\textbf{Quality at extreme head rotations.} At extreme yaw angles, artifacts can arise primarily from inaccuracies in the landmark detector and tracker, whose performance is known to degrade for non-frontal head poses. This effect may be further amplified in regions with limited training coverage. While our model remains stable under such conditions and continues to produce plausible geometry and appearance, this is a common challenge across video-based avatar methods more broadly.

\begin{figure*}[t]
\includegraphics[width=0.99\textwidth, trim=0 162cm 0 0, clip]{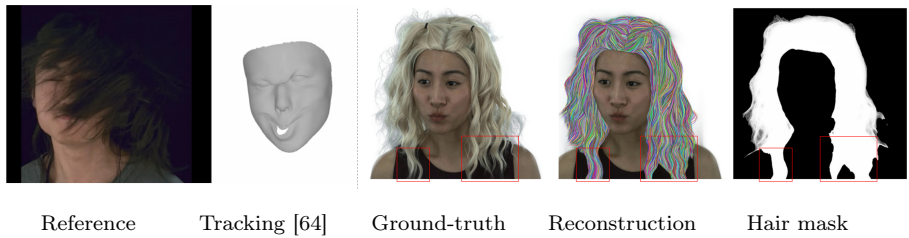}\\[2pt]
    \makebox[0.19\textwidth]{\footnotesize Reference}%
    \makebox[0.19\textwidth]{\footnotesize Tracking~\cite{xu2023gaussianheadavatar}}%
    \makebox[0.19\textwidth]{\footnotesize Ground-truth}
    \makebox[0.19\textwidth]{\footnotesize Reconstruction}%
    \makebox[0.19\textwidth]{\footnotesize Hair mask}%
    
    \caption{Limitations of our method: tracking errors of the face caused by occlusions (left) and hair structure in regions where the segmentation masks have artifacts.}
    \label{fig:limitations}
\end{figure*}

\textbf{Tracking and segmentation failures.} Our method can also be affected by loss of tracking during fast motion and by inaccuracies in the segmentation maps, which can lead to poor disentanglement between clothing and hair color, see Fig.~\ref{fig:limitations}.

\textbf{Inherited limitations from GHA.} As we build our representation for the face and upper body on the GHA method~\cite{xu2023gaussianheadavatar}, we inherit its limitations. In particular, shoulder motion is not handled well, since it relies solely on BFM tracking.

\begin{table*}[t]
\centering
\caption{Quantitative comparison on new sequences. Metrics are averaged across all views and reported separately for the novel view.}
\label{tab:comparison_new_scenes}
\setlength{\tabcolsep}{4pt}
\renewcommand{\arraystretch}{1.15}
\resizebox{\textwidth}{!}{%
\begin{tabular}{cc|l|ccc|ccc|cc|c}
\multirow{2}{*}{Scenario} & \multirow{2}{*}{Subject} & \multirow{2}{*}{Method}
& \multicolumn{3}{c|}{Full Image}
& \multicolumn{3}{c|}{Hair Region}
& \multicolumn{2}{c|}{Hair Silhouette}
& Temporal \\
& & & PSNR$\uparrow$ & SSIM$\uparrow$ & LPIPS$\downarrow$
    & PSNR$\uparrow$ & SSIM$\uparrow$ & LPIPS$\downarrow$
    & IoU$\uparrow$ & tIoU$_{\text{pred}}\uparrow$
    & tLPIPS$_{\text{excess}}\downarrow$ \\
\hline
\multirow{6}{*}{\rotatebox{90}{\scriptsize Self-reenact}}
&   \multirow{3}{*}{603} & Maya & 16.86 & 0.4912 & 0.2471 & 15.64 & 0.3864 & 0.1442 & 0.758 & 0.905 & 0.0126 \\
 &  & GHA  & 21.35 & 0.6613 & 0.2019 & 21.49 & 0.6493 & 0.1186 & \textemdash & \textemdash & -0.0188 \\
 &  & Ours & 19.77 & 0.6013 & 0.2163 & 19.42 & 0.5556 & 0.1181 & 0.820 & 0.925 & 0.0036 \\
\cline{2-12}
&   \multirow{3}{*}{604} & Maya & 13.72 & 0.6023 & 0.2483 & 11.75 & 0.4642 & 0.1339 & 0.702 & 0.881 & -0.0033 \\
 &  & GHA  & 18.30 & 0.7125 & 0.2212 & 17.86 & 0.6485 & 0.1116 & \textemdash & \textemdash & -0.0210 \\
 &  & Ours & 18.42 & 0.6978 & 0.2142 & 18.15 & 0.6183 & 0.1038 & 0.855 & 0.896 & -0.0077 \\
\hline
\multirow{6}{*}{\rotatebox{90}{\scriptsize Novel view}}
&   \multirow{3}{*}{603} & Maya & 17.72 & 0.4692 & 0.2695 & 16.70 & 0.3594 & 0.1542 & 0.784 & 0.912 & 0.0152 \\
 &  & GHA  & 21.36 & 0.6578 & 0.2224 & 22.33 & 0.6486 & 0.1253 & \textemdash & \textemdash & -0.0190 \\
 &  & Ours & 20.06 & 0.5937 & 0.2370 & 19.87 & 0.5464 & 0.1268 & 0.832 & 0.931 & 0.0051 \\
\cline{2-12}
&    \multirow{3}{*}{604} & Maya & 13.55 & 0.5813 & 0.2889 & 11.97 & 0.4417 & 0.1553 & 0.718 & 0.892 & -0.0038 \\
 &  & GHA  & 17.86 & 0.6960 & 0.2581 & 18.87 & 0.6226 & 0.1326 & \textemdash & \textemdash & -0.0254 \\
 &  & Ours & 18.06 & 0.6750 & 0.2521 & 18.59 & 0.5882 & 0.1219 & 0.871 & 0.904 & -0.0096 \\
\hline
\end{tabular}%
}
\end{table*}

\subsection{Additional discussion}

\textbf{Reliance on human-defined priors.} While our work depends on additional priors, our modeling choices are minimal and physically grounded: (i) a strand-based parametrization that enforces meaningful curve structure on hair, (ii) conditioning the motion network on standard inputs to rigid-body dynamics, such as angular velocity, acceleration, and gravity, and (iii) an elastic term that discourages non-physical stretching. The temporal dynamics themselves are fully learned from data via the LSTM and MLP. Given the limited scale of available multi-view dynamic hair data, these constraints serve to enforce physical plausibility without overly restricting what the network can learn.

\smallskip
\noindent\textbf{Frame-to-frame sequential optimization vs. canonical prediction.} Unlike methods such as HHAvatar, which represent hair with unstructured Gaussians and predict each frame from the previous hair state, DynHair predicts each frame relative to a canonical hairstyle. We hypothesize that frame-to-frame prediction yields smoother short-term motion but is more susceptible to drift accumulating over long sequences, a failure mode that canonical-frame prediction avoids by design. Each approach carries distinct tradeoffs between local smoothness and long-term stability, and combining the strengths of both is an interesting direction for future work.

\end{document}